\documentclass{article} 
\usepackage{iclr2025_conference,times}

\usepackage{amsmath,amsfonts,bm}

\def\eqref#1{equation~\ref{#1}}

\def\1{\bm{1}}

\def\rva{{\mathbf{a}}}
\def\rvb{{\mathbf{b}}}

\def\rvf{{\mathbf{f}}}
\def\rvg{{\mathbf{g}}}

\def\rvs{{\mathbf{s}}}

\def\rvv{{\mathbf{v}}}
\def\rvw{{\mathbf{w}}}
\def\rvx{{\mathbf{x}}}
\def\rvy{{\mathbf{y}}}
\def\rvz{{\mathbf{z}}}

\DeclareMathAlphabet{\mathsfit}{\encodingdefault}{\sfdefault}{m}{sl}
\SetMathAlphabet{\mathsfit}{bold}{\encodingdefault}{\sfdefault}{bx}{n}

\newcommand{\norm}[1]{\left\lVert#1\right\rVert}
\newcommand{\abs}[1]{\left|#1\right|}

\newcommand*\diff{\mathop{}\!\mathrm{d}}

\usepackage{mathrsfs}

\def\eqref#1{(\ref{#1})}

\def\eqref#1{(\ref{#1})}

\definecolor{brightmaroon}{rgb}{0.76, 0.13, 0.28}

\def\eqref#1{(\ref{#1})}

\usepackage[shortlabels]{enumitem}
\usepackage{minitoc}
\usepackage[nottoc]{tocbibind}

\usepackage[toc,page,header]{appendix}

\usepackage{hyperref}
\usepackage{multirow}
\usepackage{url}
\usepackage{mathabx}
\usepackage{graphicx}
\usepackage{wrapfig}
\usepackage{subcaption}
\usepackage{float}
\usepackage{algorithm}
\usepackage{algpseudocode}
\algrenewcommand\algorithmicindent{0.2em}
\usepackage{fancybox}
\usepackage[framemethod=TikZ]{mdframed}
\mdfsetup{%
	middlelinecolor=black,
	middlelinewidth=1.2pt,
	backgroundcolor=gray!10,
	roundcorner=10pt}
\usepackage{amsthm}
\usepackage{cleveref}
\theoremstyle{plain}
\newtheorem{theorem}{Theorem}[section]

\newtheorem{proposition}{Proposition}[section]

\theoremstyle{definition}
\newtheorem{assumption}{Assumption}[section]
\newtheorem{definition}{Definition}[section]

\theoremstyle{remark}

\title{Bellman Diffusion: Generative Modeling as Learning a Linear Operator in the Distribution Space}

\author{Yangming Li$^{1,}$\thanks{Joint first authors.} , Chieh-Hsin Lai$^{2,*}$, Carola-Bibiane Schönlieb$^1$, Yuki Mitsufuji$^2$, \& Stefano Ermon$^3$ \\
	$^1$Department of Applied Mathematics and Theoretical Physics, University of Cambridge \\
	$^2$Sony AI \\
	$^3$Department of Computer Science, Stanford University \\
	\texttt{yl874@cam.ac.uk,chieh-hsin.lai@sony.com}
}

\iclrfinalcopy 
\begin{document}
	\doparttoc 
	\faketableofcontents 
	
	\maketitle
	
	\begin{abstract}
		
		Deep Generative Models (DGMs), including Energy-Based Models (EBMs) and Score-based Generative Models (SGMs), have advanced high-fidelity data generation and complex continuous distribution approximation. However, their application in Markov Decision Processes (MDPs), particularly in distributional Reinforcement Learning (RL), remains underexplored, with the classical histogram-based methods dominating the field. This paper rigorously highlights that this application gap is caused by the nonlinearity of modern DGMs, which conflicts with the linearity required by the Bellman equation in MDPs. For instance, EBMs involve nonlinear operations such as exponentiating energy functions and normalizing constants. To address this problem, we introduce \emph{Bellman Diffusion}, a novel DGM framework that maintains linearity in MDPs through gradient and scalar field modeling. With divergence-based training techniques to optimize neural network proxies and a new type of  stochastic differential equation (SDE) for sampling, Bellman Diffusion is guaranteed to converge to the target distribution. Our empirical results show that Bellman Diffusion achieves accurate field estimations and is a capable image generator, converging $1.5 \times$ faster than the traditional histogram-based baseline in distributional RL tasks. This work enables the effective integration of DGMs into MDP applications, unlocking new avenues for advanced decision-making frameworks.
		
	\end{abstract}
	
	\section{Introduction}\label{sec: intro}
	
	Even though Deep Generative Models (DGMs), such as Energy-Based Models (EBMs)~\citep{teh2003energy}, Generative Adversarial Networks (GANs)~\citep{goodfellow2020generative}, and the emerging Score-based Generative Models (SGMs)~\citep{song2021scorebased,sohl2015deep,ho2020denoising}, have been highly developed to achieve high-fidelity generation~\citep{li2023your,wang2023diffusion,krishnamoorthy2023diffusion} and accurately approximate complex continuous distributions, their applications in Markov Decision Processes (MDPs), including Planning and Reinforcement Learning (RL), remain underexplored. Instead, a classical histogram-based methods (e.g., C51~\citep{bellemare2017distributional}) are still widely used in MDPs. We identify the main reason for this gap within the core of MDPs: the Bellman equation~\citep{bellemare2017distributional}, which supports efficient model training but imposes a linear structure that conflicts with the strong nonlinearity of modern DGMs. In the following, we present a case study that shows this bottleneck using EBMs.

	\paragraph{Motivation and problem.} The easiest way to model the return distribution with EBMs is to sample full state-action trajectories, so that one can directly use the returns computed from trajectories to train EBMs. However, this is not scalable since trajectory sampling is very costly in many RL environments. Some such examples  are provided in Appendix~\ref{sec:apply to mdp}. An alternative is to follow the paradigm of Deep Q-Learning~\citep{mnih2013playing}, which applies the distributional Bellman equation to update the model with only partial trajectories. This equation is linear in form, which connects the return distribution $p_z(x)$ of a state $z$ with that of the next state $z'$, expressed formally as:
	\begin{equation}
		\label{eq:bellman example}
		p_{z}(x) = \sum_{z', r} \alpha_{z, z', r} p_{z'}\Big(\frac{x - r}{\gamma}\Big),
	\end{equation}
	where $r$ is the expected reward for transitioning between states, and $\alpha_{z, z', r}$, $\gamma$ are constants determined by the RL environment. Although EBMs (or existing DGMs) are effective in distribution approximation~\citep{lee2023convergence}, their application to the Bellman equation is constrained by the nonlinearity of how it is modeling
	$\frac{e^{-E_z (x)}}{\mathcal{Z}_z}$, where $ \mathcal{Z}_z$ is the normalization factor at each state $z$. To simply the case, suppose that the term $\mathcal{Z}_z$ is always $1$ for every state $z$, then we have:
	\begin{equation}\label{eq:nonlinear_dsm}
		\begin{aligned}
			E_z (x) & = - \log (\mathcal{Z}_z p_{z}(x)) = -\log \Big( \sum_{z', r} \alpha_{z, z', r} p_{z'}\Big(\frac{x - r}{\gamma}\Big) \Big) \\
			& \le - \sum_{z', r} \alpha_{z, z', r} \log p_{z'}\Big(\frac{x - r}{\gamma}\Big) = \sum_{z', r} \alpha_{z, z', r} E_z \Big(\frac{x - r}{\gamma}\Big)
		\end{aligned},
	\end{equation}
	which holds due to Jensen's inequality, indicating that the energy function $E_z(\mathbf{x})$ for a state $z$ is not linearly interpolated by the energy function for the next state $z'$. As such, EBMs act as a nonlinear operator, disrupting the linearity of the distributional Bellman equation, thereby rendering EBMs inapplicable in this context. In Sec.~\ref{sec:prelim}, we analyze the modeling approaches of other modern DGMs and find that none can preserve the linearity of the Bellman update.
	
	\paragraph{Our framework: Bellman Diffusion.} 
	In this work, we re-examine the modeling approach of well-known DGMs to assess whether they meet the desired linearity property (Sec.~\ref{sec:prelim}), revealing that none of the existing DGMs satisfy this requirement. To address this bottleneck, we propose \emph{Bellman Diffusion}, a novel DGM designed to overcome bottlenecks in applying DGMs to MDPs. The core idea is to model the gradient field $\nabla p_{z}(x)$ and scalar field $p_{z}(x)$ directly. Since these fields are linear operators, the linearity of Bellman equations is well preserved. For instance, after applying the gradient operator $\nabla$, the Bellman equation remains linear:
	\begin{equation}
		\nabla p_{z}(x) = \sum_{z', r} \frac{\alpha_{z, z', r}}{\gamma} \nabla p_{z'}\Big(\frac{x - r}{\gamma}\Big).
	\end{equation}
	We now use $p_{\mathrm{target}}$ to denote the target density of each state, replacing the previous notation $p_z$. Since $\nabla p_{\mathrm{target}}(\rvx)$ and $p_{\mathrm{target}}(\rvx)$ are generally inaccessible, we introduce field-based divergence measures (Definition~\ref{def: field div}) and transform them into training objectives (Eqs.~\eqref{eq: final version of gradient loss}, \eqref{eq: final version of scalar loss}): approximating fields $\nabla p_{\mathrm{target}}(\rvx)$ and $p_{\mathrm{target}}(\rvx)$ with neural network proxies $\rvg_{\bm{\phi}}$ and $s_{\bm{\varphi}}$.
	
	Given these proxies, we introduce a new sampling method: Bellman Diffusion Dynamics, associated with the fields represented by the following stochastic differential equation (SDE):
	\begin{equation}\label{eq:intro_sde_sampling}
		\diff \rvx(t) = \underbrace{\nabla p_{\mathrm{target}}(\rvx(t))}_{\approx \rvg_{\bm{\phi}}} \diff t + \underbrace{\sqrt{p_{\mathrm{target}}(\rvx(t))}}_{\approx \sqrt{s_{\bm{\varphi}}}} \diff\rvw(t), \quad \text{starting from } \rvx(0)\sim p_0,
	\end{equation}
	where $\rvw(t)$ is a Brownian process and $p_0$ is any initial distribution. Once the fields are well approximated, we can replace the field terms in the above equation with learned proxies, resulting in a proxy SDE that can be solved forward in time to sample from $p_{\mathrm{target}}(\rvx)$.
	
	\paragraph{Theoretical and empirical results.} Theoretically, we guarantee the convergence of our Bellman Diffusion Dynamics to the stationary distribution $p_{\mathrm{target}}$, regardless of the initial distribution (Theorem~\ref{thm:stationary_dist}), and provide an error bound analysis accounting for neural network approximation errors (Theorem~\ref{thm:informal_error}). Thus, Bellman Diffusion is a reliable standalone generative model. 
	
	Experimentally, we show the generative capabilities of Bellman Diffusion on real and synthetic datasets, confirming accurate field estimations, with promising results in image generation. We further apply Bellman Diffusion to classical distributional RL tasks, resulting in much more stable and $1.5 \times$ faster convergence compared to the widely used histogram method~\citep{bellemare2017distributional}. Notably, it can effectively learn and recover the target distributions with multiple unbalanced modes (Fig.~\ref{fig:toy middle results} of Sec.~\ref{sec:experiments}), a challenge for score-based methods~\citep{song2019generative} due to the inherent nature of the score function.
	
	In summary, Bellman Diffusion introduced in this paper stands as a novel and mathematically grounded generative modeling approach, paving the way for continuous density modeling in various applications within MDPs, such as Planning and distributional RL.

	\section{Linear Property for MDPs}\label{sec:prelim}
	
	In this section, we review modern DGMs and highlight the desired property to facilitate density estimation with Bellman updates, avoiding full trajectory updates.
	
	\subsection{Modelings of Modern Deep Generative Models}
	\label{sec:current dgms}
	
	DGMs aim to model the complex target distribution $p_{\mathrm{target}}(\mathbf{x})$ using a neural network-approximated \emph{continuous} density, enabling new samples generation. 
	Below, we review well-known DGMs and offer high-level insights into how they define a \emph{modeling operator} $\mathcal M$ that connects their own modeling functions to the desired density or its related statistics.
	
	\textbf{Energy-Based Models (EBMs)~\citep{teh2003energy}:} These models define an energy function $ E(\mathbf{x}) $ and represent the probability as: $p_{\mathrm{target}}(\mathbf{x}) \approx \frac{e^{-E(\mathbf{x})}}{Z}$, where $ Z :=\int e^{-E(\mathbf{x})} \diff\rvx$  is the partition function for normalizing probabilities. EBM defines a modeling operator $\mathcal M_{\mathrm{EBM}}\colon E(\cdot) \mapsto \frac{e^{-E(\cdot)}}{Z}$, linking the statistic $E(\cdot)$ to desired density.
	
	\textbf{Flow-Based Models~\citep{rezende2015variational,chen2018neural}:} These use a series of invertible transformations $ \rvf(\rvx) $ to map data $\mathbf{x}$ to a latent space $\mathbf{z}$, with an exact likelihood: $p_{\mathrm{target}}(\mathbf{x}) \approx \pi(\mathbf{z}) \left| \det \frac{\partial \rvf^{-1}(\rvx)}{\partial \mathbf{x}} \right|$. It determines a modeling operator $\mathcal M_{\mathrm{Flow}}\colon \rvf(\cdot) \mapsto \pi(\rvf(\cdot)) \left| \det \frac{\partial \rvf^{-1}(\cdot)}{\partial \mathbf{x}} \right|$, connecting the transformation $\rvf(\cdot)$ to desired density.

	\textbf{Implicit Latent Variable Models:} These models define a latent variable $ \mathbf{z} $ and use a generative process $ p(\mathbf{x}|\mathbf{z}) $, where the latent space is sampled from a prior $ \pi(\mathbf{z}) $, usually taken as a standard normal distribution. Two popular models are VAE~\citep{kingma2013auto} and GAN~\citep{goodfellow2020generative}. VAE maximizes a variational lower bound using an encoder network $ q(\mathbf{z}|\mathbf{x}) $ to approximate the posterior distribution, while GAN employs a discriminator to distinguish between real and generated data, with a generator learning to produce realistic samples but lacking an explicit likelihood.
	
	Since VAEs and GANs are implicit models, they lack an explicit modeling operator like $\mathcal M_{\mathrm{EBM}}$ and $\mathcal M_{\mathrm{Flow}}$ that connects modeling functions to the desired density or its related statistics.
	
	\textbf{Score-Based Generative Models (SGMs)~\citep{song2021scorebased}:} They involve a process that gradually adds noise to $ p_{\mathrm{target}} $, resulting in a sequence of time-conditioned densities $ \{p(\rvx_t, t)\}_{t \in [0,T]} $, where $ t = 0 $ corresponds to $ p_{\mathrm{target}} $ and $ t = T $ corresponds to a simple prior distribution $ \pi(\rvz) $. Then, SGMs reverse this diffusion process for sampling by employing the time-conditioned score $ \rvs(\cdot,t) := \nabla \log p(\cdot, t) $ and solving the ordinary differential equation~\citep{song2021scorebased} from $t=T$ to $t=0$ with $ \bm{\phi}_T(\rvx_T) = \rvx_T \sim \pi $:
	$
	\diff\bm{\Psi}_t(\rvx_T) = \left(\rvf(\bm{\Psi}_t(\rvx_T), t) - \frac{1}{2}g^2(t) \rvs(\bm{\Psi}_t(\rvx_T), t) \right)\diff t$, where $\rvf$ and $g$ are pre-determined. This flow defines a pushforward map $ \mathcal{V}^{T\rightarrow t}[\rvs] $ of the density as
	$
	\mathcal{V}^{T\rightarrow t}[\rvs]\{\pi\} := \pi\left(\bm{\Psi}_t^{-1}(\cdot)\right) \left| \det \frac{\partial \bm{\Psi}_t^{-1}(\cdot)}{\partial \mathbf{x}} \right|
	$. Thus, SGMs determine a modeling operator $ \mathcal{M}_{\mathrm{SGM}} \colon \rvs \mapsto \mathcal{V}^{T\rightarrow 0}[\rvs]  \{\pi\} \approx p_{\mathrm{target}} $.
	
	\subsection{Desired Linear Property in MDP}
	As the case of EBMs shown in Sec.~\ref{sec: intro}, to leverage the strong capability of DGMs in density modeling with the Bellman update (Eq.~\eqref{eq:bellman example}), the linearity of modeling operator $\mathcal{M}$ is crucial:
	\begin{mdframed}
		\textit{ \textbf{Linear property of modeling.} The modeling operator $\mathcal{M}$ defined by a DGM is linear: $\mathcal{M}\big(a f + b g\big) =a \mathcal{M}\big(f\big) + b \mathcal{M}\big( g\big)$, for any reals $a, b$ and functions $f, g$.}
	\end{mdframed}
	If $\mathcal{M}$ is linear, we can link future state densities or their statistics with the current state for efficient updates, as shown in the Bellman equation in Eq.~\eqref{eq:bellman example}:
	\begin{equation*} 
		\mathcal{M}(p_{z})(x) = \sum_{z', r} \alpha_{z, z', r} \mathcal{M}(p_{z'})\left(\frac{x - r}{\gamma}\right). 
	\end{equation*}
	However, for the current well-established DGMs, their modeling operators are either not explicitly defined (e.g., VAE and GAN), thus lacking guaranteed linearity, or are nonlinear operators (e.g., $\mathcal{M}_{\mathrm{EBM}}$, $\mathcal{M}_{\mathrm{Flow}}$, and $\mathcal{M}_{\mathrm{SGM}}$  as shown in Ineq.~\eqref{eq:nonlinear_dsm}). Consequently, this restricts the application of these powerful DGMs to MDPs.
	
	\section{Method: Bellman Diffusion}
	\label{sec:method overview}
	
	In this section, we mainly provide an overview of Bellman Diffusion, presenting the usage, with its theoretical details later in Sec.~\ref{sec:theory}. We defer all proofs to Appendix~\ref{sec:method overview}.
	
	\subsection{Scalar and Vector Field Matching}
	\label{sec:field_matching}
	
	\paragraph{Field matching.} Suppose we have a finite set of $D$-dimensional sample vectors $\mathcal{X} = \{ \mathbf{x}_i \}_{1 \le i \le N}$, where each point $\mathbf{x}_i$ is drawn from the distribution ${p_{\mathrm{target}}(\rvx)}$. Bellman Diffusion, as a generative model, aims to learn both the gradient field $\nabla {p_{\mathrm{target}}(\rvx)}$ and the scalar field ${p_{\mathrm{target}}(\rvx)}$. Like Fisher divergence~\citep{antolin2009fisher} for the score function $\nabla \log {p_{\mathrm{target}}(\rvx)}$, we introduce two divergence measures for $\nabla {p_{\mathrm{target}}(\rvx)}$ and ${p_{\mathrm{target}}(\rvx)}$.
	
	\begin{definition}[Field Divergences]\label{def: field div}
		Let $p(\cdot)$ and $q(\cdot)$ be continuous probability densities. The discrepancy between the two can be defined as 
		\begin{equation} \label{eq:def of grad div}
			\mathcal{D}_{\mathrm{grad}}\big(p(\cdot), q(\cdot)\big) = \int p(\mathbf{x}) \norm{ \nabla p(\mathbf{x}) - \nabla q(\mathbf{x}) }^2 \diff \rvx 
		\end{equation} using the gradient operator $\nabla$ in terms of $\rvx$, or as 
		\begin{equation} \label{eq:def of scalar div}
			\mathcal{D}_{\mathrm{id}}\big(p(\cdot), q(\cdot)\big) = \int p(\mathbf{x}) ( p(\mathbf{x}) - q(\mathbf{x}) )^2 \diff\rvx 
		\end{equation} using the identity operator $\mathbb{I}$. Here, $\norm{\cdot}$ denotes the $\ell_2$ norm.
	\end{definition}
	
	As shown in Appendix~\ref{sec:validity_field_divergence}, the two measures above are valid statistical measures. These measures are used to empirically estimate the gradient field $\nabla {p_{\mathrm{target}}(\rvx)}$ and the scalar field ${p_{\mathrm{target}}(\rvx)}$ from real data $\mathcal{X}$. . Furthermore, our modeling defines a modeling operator given by $\mathcal{M}_{\mathrm{Bellman}}:=\begin{bmatrix}
		\nabla \\
		\mathbb{I}
	\end{bmatrix}\colon p_{\mathrm{target}}(\cdot) \mapsto \begin{bmatrix}
		\nabla p_{\mathrm{target}}(\cdot)  \\
		p_{\mathrm{target}}(\cdot) 
	\end{bmatrix}$ which is linear in its input. 
	
	Similar to SGMs, we parameterize two neural networks, $\mathbf{g}_{\bm{\phi}}(\mathbf{x})$ and $s_{\bm{\varphi}}(\mathbf{x})\geq 0$, to approximate these fields, with learnable parameters $\bm{\phi}$ and $\bm{\varphi}$, using the following estimation loss functions:
	\begin{equation}
		\label{eq:initial loss forms}
		\left\{\begin{aligned}
			\mathcal{L}_{\mathrm{grad}}(\bm{\phi}) & := \mathcal{D}_{\mathrm{grad}}\big({p_{\mathrm{target}}(\cdot)}, \mathbf{g}_{\bm{\phi}}(\cdot)\big) = \mathbb{E}_{\mathbf{x} \sim {p_{\mathrm{target}}(\rvx)}} \Big[ \| \nabla {p_{\mathrm{target}}(\rvx)} - \mathbf{g}_{\bm{\phi}}(\mathbf{x})  \|^2 \Big] \\
			\mathcal{L}_{\mathrm{id}}(\bm{\varphi}) & := \mathcal{D}_{\mathrm{id}}\big({p_{\mathrm{target}}(\cdot)}, s_{\bm{\varphi}} (\cdot)\big) = \mathbb{E}_{\mathbf{x} \sim {p_{\mathrm{target}}(\rvx)}} \Big[ ( {p_{\mathrm{target}}(\rvx)} - s_{\bm{\varphi}}(\mathbf{x}) )^2 \Big]
		\end{aligned}\right.
	\end{equation}
	Since the terms $\nabla {p_{\mathrm{target}}(\rvx)}$ and ${p_{\mathrm{target}}(\rvx)}$ inside the expectation are generally inaccessible, these losses cannot be estimated via Monte Carlo sampling. The following proposition  resolves this issue by deriving a feasible proxy for the loss functions
	\begin{proposition}[Equivalent Forms of Field Matching]
		\label{prop:reshaped loss}
		The loss $\mathcal{L}_{\mathrm{grad}}(\bm{\phi})$ is given by
		\begin{equation*}
			\mathcal{L}_{\mathrm{grad}}(\bm{\phi}) = C_{\mathrm{grad}} + \lim_{\epsilon \rightarrow 0} \mathbb{E}_{\mathbf{x}_1, \mathbf{x}_2 \sim {p_{\mathrm{target}}(\rvx)}} \Big[ \| \mathbf{g}_{\bm{\phi}}(\mathbf{x}_1) \|^2 + \mathrm{tr}( \nabla \mathbf{g}_{\bm{\phi}}(\mathbf{x}_1))  \mathcal{N}(\mathbf{x}_2 - \mathbf{x}_1; \mathbf{0}, \epsilon \mathbf{I}_D )  \Big],
		\end{equation*}
		and $\mathcal{L}_{\mathrm{id}}(\bm{\varphi})$ is expressed as
		\begin{equation*}
			\mathcal{L}_{\mathrm{id}}(\bm{\varphi}) = C_{\mathrm{id}} + \lim_{\epsilon \rightarrow 0} \mathbb{E}_{\mathbf{x}_1, \mathbf{x}_2 \sim {p_{\mathrm{target}}(\rvx)}}\Big[s_{\bm{\varphi}}(\mathbf{x}_1)^2  - 2 s_{\bm{\varphi}}(\mathbf{x}_1) \mathcal{N}(\mathbf{x}_2 - \mathbf{x}_1; \mathbf{0}, \epsilon \mathbf{I}_D)\Big].
		\end{equation*}
		Here, $\mathcal{N}(\rvx; \mathbf{0}, \epsilon \mathbf{I}_D)$ denotes a $D$-dimensional isotropic Gaussian density function with $\rvx$, and $C_{\mathrm{grad}}$ and $C_{\mathrm{id}}$ are constants independent of the model parameters $\bm{\phi}$ and $\bm{\varphi}$.
	\end{proposition}
	Building on the above proposition, we can obtain feasible approximations of the training losses. With $\epsilon$ fixed to be sufficiently small (see Sec.~\ref{sec:experiments} for experimental setups), we have:
	\begin{gather}
		\label{eq:relaxed loss}
		\left\{\begin{aligned}
			\widebar{\mathcal{L}}_{\mathrm{grad}}(\bm{\phi};\epsilon) &:= C_{\mathrm{grad}} + \mathbb{E}_{\mathbf{x}_1, \mathbf{x}_2 \sim {p_{\mathrm{target}}(\rvx)}} \Big[ \| \mathbf{g}_{\bm{\phi}}(\mathbf{x}_1) \|^2 + \mathrm{tr}( \nabla \mathbf{g}_{\bm{\phi}}(\mathbf{x}_1))  \mathcal{N}(\cdot, \mathbf{0}, \epsilon \mathbf{I}_D )  \Big] \approx \mathcal{L}_{\mathrm{grad}}(\bm{\phi}) ,\\
			\widebar{\mathcal{L}}_{\mathrm{id}}(\bm{\varphi};\epsilon) &:= C_{\mathrm{id}} + \mathbb{E}_{\mathbf{x}_1, \mathbf{x}_2 \sim {p_{\mathrm{target}}(\rvx)}}\Big[s_{\bm{\varphi}}(\mathbf{x}_1)^2  - 2 s_{\bm{\varphi}}(\mathbf{x}_1) \mathcal{N}(\mathbf{x}_2 - \mathbf{x}_1, \mathbf{0}, \epsilon \mathbf{I}_D)\Big] \approx  \mathcal{L}_{\mathrm{id}}(\bm{\varphi})
		\end{aligned}\right.
	\end{gather}
	We note that scalar and gradient fields can be modeled independently.
	Moreover, as Bellman Diffusion directly matches these fields, it eliminates the need for the normalizing constant associated with costly spatial integrals in the density network required by EBMs.
	
	\subsection{Efficient Field Matching Losses}
	
	\paragraph{Slice trick for efficient training.}
	
	While the loss functions $\widebar{\mathcal{L}}_{\mathrm{grad}}(\bm{\phi};\epsilon)$ and $\widebar{\mathcal{L}}_{\mathrm{id}}(\bm{\varphi};\epsilon)$ support Monte Carlo estimation, the term $\mathrm{tr}(\nabla \mathbf{g}_{\bm{\phi}}(\mathbf{x}_1))$ in $\widebar{\mathcal{L}}_{\mathrm{grad}}(\bm{\phi};\epsilon)$ is computationally expensive, limiting the scalability in high dimensions. To address this problem, we apply the slice trick~\citep{kolouri2019generalized,song2020sliced} to estimate the trace term efficiently. The resulting objective is summarized in the following proposition.
	\begin{proposition}[Sliced Gradient Matching]
		\label{prop:sliced gradient matching}
		We define the sliced version of  $\mathcal{L}_{\mathrm{grad}}$ (i.e., Eq.~(\ref{eq:def of grad div})) as
		\begin{equation*}
			\mathcal{L}_{\mathrm{grad}}^{\mathrm{slice}}(\bm{\phi}) = \mathbb{E}_{\mathbf{v} \sim q(\mathbf{v}), \mathbf{x} \sim {p_{\mathrm{target}}(\rvx)}} \Big[ \big( \mathbf{v}^T \nabla {p_{\mathrm{target}}(\rvx)} - \mathbf{v}^T \mathbf{g}_{\bm{\phi}}(\mathbf{x})  \big)^2 \Big],
		\end{equation*}    
		where $\mathbf{v}$ represents the slice vector drawn from a continuous distribution $q(\mathbf{v})$. This sliced loss also has an equivalent form: 
		\begin{equation*}
			\mathcal{L}_{\mathrm{grad}}^{\mathrm{slice}} (\bm{\phi}) = C_{\mathrm{grad}}' + \lim_{\epsilon \rightarrow 0} \mathbb{E}_{\substack{\mathbf{v} \sim q(\mathbf{v}); \\ \mathbf{x}_1, \mathbf{x}_2 \sim {p_{\mathrm{target}}(\rvx)}}} \Big[ (\mathbf{v}^{\top} \mathbf{g}_{\bm{\phi}}(\mathbf{x}_1))^2 + ( \mathbf{v}^{\top} \nabla_{\mathbf{x}_1} \mathbf{g}_{\bm{\phi}}(\mathbf{x}_1) \mathbf{v} ) \mathcal{N}(\mathbf{x}_2 -\mathbf{x}_1; \mathbf{0}, \epsilon \mathbf{I}_D) \Big],
		\end{equation*}
		where $C_{\mathrm{grad}}'$ is another constant independent of the model parameters.
	\end{proposition}

	Similar to Eq.~\eqref{eq:relaxed loss}, we can define a proxy loss for $\mathcal{L}_{\mathrm{grad}}^{\mathrm{slice}} (\bm{\phi})$ as follows with a sufficiently small $\epsilon$:
	\begin{equation}
		\label{eq: continuously relaxed gradient loss}
		\mathbb{E}_{\substack{\mathbf{v} \sim q(\mathbf{v});  \mathbf{x}_1, \mathbf{x}_2 \sim {p_{\mathrm{target}}(\rvx)}}}  \Big[ (\mathbf{v}^{\top} \mathbf{g}_{\bm{\phi}}(\mathbf{x}_1))^2 + ( \mathbf{v}^{\top} \nabla_{\mathbf{x}_1} \mathbf{g}_{\bm{\phi}}(\mathbf{x}_1) \mathbf{v} ) \mathcal{N}(\mathbf{x}_2 -\mathbf{x}_1; \mathbf{0}, \epsilon \mathbf{I}_D) \Big],
	\end{equation}
	which allows Monte Carlo estimation from samples $\mathcal{X}$. This proxy loss serves as a reasonable estimator~\citep{lai2023fp} for $\mathcal{L}_{\mathrm{grad}}(\bm{\phi})$.

	\paragraph{Slice trick for improving sample efficiency.}  
	
	When the data dimension $D$ is large, the multiplier $\mathcal{N}(\mathbf{x}_2 - \mathbf{x}_1; \mathbf{0}, \epsilon \mathbf{I}_D)$ in the loss functions: $\widebar{\mathcal{L}}_{\mathrm{grad}}(\bm{\phi};\epsilon)$ and $\widebar{\mathcal{L}}_{\mathrm{id}}(\bm{\varphi};\epsilon)$, will become nearly zero due to the $(2\pi)^{-D/2}$ factor, requiring a very large batch size for accurate Monte Carlo estimation and leading to low data efficiency.
	
	To resolve this issue, we apply an additional slice trick, projecting the $D$-dimensional Gaussian density $\mathcal{N}(\mathbf{x}_2 - \mathbf{x}_1; \mathbf{0}, \epsilon \mathbf{I}_D)$ into a 1-dimensional density $\mathcal{N}(\mathbf{w}^T \mathbf{x}_2 - \mathbf{w}^T \mathbf{x}_1, 0, \epsilon)$ along a random direction $\mathbf{w}\sim q(\rvw)$, where $\mathbf{w}$ follows a slice vector distribution $q(\rvw)$. Combining with Eq.~\eqref{eq: continuously relaxed gradient loss}, this results in our ultimate gradient field matching loss:
	\begin{mdframed}
		\begin{equation}
			\label{eq: final version of gradient loss}
			\widebar{\mathcal{L}}_{\mathrm{grad}}^{\mathrm{slice}}(\bm{\phi};\epsilon) := \mathbb{E}_{\substack{\mathbf{w} \sim q(\mathbf{w}), \mathbf{v} \sim q(\mathbf{v});\\ \mathbf{x}_1, \mathbf{x}_2 \sim {p_{\mathrm{target}}(\rvx)}}} \Big[ (\mathbf{v}^{\top} \mathbf{g}_{\bm{\phi}}(\mathbf{x}_1))^2 + ( \mathbf{v}^{\top} \nabla_{\mathbf{x}_1} \mathbf{g}_{\bm{\phi}}(\mathbf{x}_1) \mathbf{v} ) \mathcal{N}(\mathbf{w}^T\mathbf{x}_2 - \mathbf{w}^T\mathbf{x}_1; 0, \epsilon) \Big].
		\end{equation}
	\end{mdframed} 
	Similarly, we apply the same trick to $\widebar{\mathcal{L}}_{\mathrm{id}}(\bm{\varphi};\epsilon)$ for dimension projection and obtain:
	\begin{mdframed}
		\begin{equation}
			\label{eq: final version of scalar loss}
			\widebar{\mathcal{L}}_{\mathrm{id}}^{\mathrm{slice}}(\bm{\varphi};\epsilon) := \mathbb{E}_{\substack{\mathbf{w} \sim q(\mathbf{w});\\ \mathbf{x}_1, \mathbf{x}_2 \sim {p_{\mathrm{target}}(\rvx)}}}\Big[s_{\bm{\varphi}}(\mathbf{x}_1)^2  - 2 s_{\bm{\varphi}}(\mathbf{x}_1) \mathcal{N}(\mathbf{w}^T\mathbf{x}_2 - \mathbf{w}^T\mathbf{x}_1; 0, \epsilon)\Big].
		\end{equation}
	\end{mdframed} 
	We adopt $\widebar{\mathcal{L}}_{\mathrm{grad}}^{\mathrm{slice}}(\bm{\phi};\epsilon)$ and $\widebar{\mathcal{L}}_{\mathrm{id}}^{\mathrm{slice}}(\bm{\varphi};\epsilon)$ for vector and scalar field matching losses, as they offer more practical and efficient objectives than $\mathcal{L}_{\mathrm{grad}}(\bm{\phi})$ and $\mathcal{L}_{\mathrm{id}}(\bm{\varphi})$, respectively. Empirically, these adaptations significantly stabilize the model in experiments. 
	
	\subsection{Bellman Diffusion Dynamics}
	
	Suppose that neural networks $\mathbf{g}_{\bm{\phi}}(\mathbf{x}), s_{\bm{\varphi}}(\mathbf{x})$ accurately estimate the target fields $\nabla p_{\mathrm{target}}(\rvx)$ and $p_{\mathrm{target}}(\rvx)$, one can sample from $p_{\mathrm{target}}(\rvx)$ by approximating the score function as:
	\begin{equation*}
		\nabla \log {p_{\mathrm{target}}(\rvx)} = \frac{\nabla {p_{\mathrm{target}}(\rvx)}}{{p_{\mathrm{target}}(\rvx)}} \approx \frac{\mathbf{g}_{\bm{\phi}}(\mathbf{x})} {s_{\bm{\varphi}}(\mathbf{x})}
	\end{equation*}
	and then applying Langevin dynamics~\citep{bussi2007accurate}:    \begin{equation*}
		\diff\mathbf{x}(t) = \nabla \log {p_{\mathrm{target}}(\rvx)}  \diff t + \sqrt{2} \diff\bm{\omega}(t) \approx \frac{\mathbf{g}_{\bm{\phi}}(\mathbf{x})} {s_{\bm{\varphi}}(\mathbf{x})} \diff t + \sqrt{2} \diff\bm{\omega}(t),
	\end{equation*}
	where $\bm{\omega}(t)$ is a standard Brownian motion. However, this approach can be numerically unstable due to the division\footnote{For example, if $s_{\bm{\varphi}}(\mathbf{x})$ is around 0.01, its inverse can magnify the estimation error of $\mathbf{g}_{\bm{\phi}}(\mathbf{x})$ by 100 times.}. This issue is unavoidable as $p_{\mathrm{target}}(\rvx)$ vanishes when $\norm{\mathbf{x}} \rightarrow \infty$. Additionally, it doesn't support the distributional Bellman update for MDPs as mentioned in Sec.~\ref{sec: intro}. To solve this, we propose a new SDE for sampling from $p_{\mathrm{target}}(\rvx)$, called \emph{Bellman Diffusion Dynamics}: 
	\begin{mdframed}
		\begin{equation}\label{eq:new sampling method}
			\diff\mathbf{x}(t) = \nabla p_{\mathrm{target}}(\mathbf{x}(t)) \diff t + \sqrt{p_{\mathrm{target}}(\mathbf{x}(t))} \diff\bm{\omega}(t).
		\end{equation}
	\end{mdframed}
	We also provide the theoretical motivation and derivation of Eq.~\eqref{eq:new sampling method} in Appendix~\ref{sec:motivation}.
	
	In practice, once the neural network approximations $\mathbf{g}_{\bm{\phi}}(\mathbf{x}) \approx \nabla p_{\mathrm{target}}(\mathbf{x}) $ and $s_{\bm{\varphi}}(\mathbf{x}) \approx p_{\mathrm{target}}(\mathbf{x})  $ are both well-learned, we can derive the following \emph{empirical Bellman Diffusion Dynamics}, a feasible proxy SDE for Eq.~\eqref{eq:new sampling method}:
	\begin{equation}\label{eq:nn new sampling method}
		\diff\mathbf{x}(t) = \mathbf{g}_{\bm{\phi}}(\mathbf{x}) \diff t + \sqrt{s_{\bm{\varphi}}(\mathbf{x})} \diff\bm{\omega}(t).
	\end{equation}
	Bellman Diffusion learns and samples using both the scalar and gradient fields, allowing it to better approximate low-density regions and unbalanced target weights—unlike SGMs, which relies solely on the score (i.e., gradient-log density)~\citep{song2019generative} and cannot recognize weighted modes in simple mixture data; see \Cref{thm:informal_error} and Sec.~\ref{sec:experiments-toy}.
	
	In Sec.~\ref{sec:theory}, we also provide a steady-state analysis of Eq.~\eqref{eq:new sampling method} and an error analysis for Eq.~\eqref{eq:nn new sampling method}, supporting the rationale behind our Bellman Diffusion Dynamics.

	\subsection{Summary of Training and Sampling Algorithms}
	To summarize Bellman Diffusion as a deep generative model, we outline the training and sampling steps in Algorithm~\ref{alg:training} and Algorithm~\ref{alg:sampling}. For training, we first sample real data $\mathbf{x}_1, \mathbf{x}_2$ from dataset $\mathcal{X}$ (line 2) and slice vectors $\mathbf{v}, \mathbf{w}$ from some predefined distributions $q(\mathbf{v}), q(\mathbf{w})$ (line 3)\footnote{Here, we follow the practice in \citet{song2020sliced} by using a single slice vector to approximate the expectation over $q(\rvv)$ or $q(\rvw)$, trading variance for reduced computational cost.}.  Then, we estimate the loss functions $\widebar{\mathcal{L}}_{\mathrm{grad}}^{\mathrm{slice}}(\bm{\phi};\epsilon)$, $\widebar{\mathcal{L}}_{\mathrm{id}}^{\mathrm{slice}}(\bm{\varphi};\epsilon)$ using Monte Carlo sampling (lines 4-6). Finally, the model parameters $\bm{\phi}$ and $\bm{\varphi}$ are updated via gradient descent (lines 7-8).
	
	For inference, we begin by sampling $\mathbf{x}(0)$ from an arbitrary distribution, such as standard normal (line 1). Then, after setting the number of steps $T$ and step size $\eta$, we iteratively update $\mathbf{x}(0)$ to $\mathbf{x}(\eta T)$ following Eq.~\eqref{eq:new sampling method} (lines 3-7).
	
	\begin{figure}[t]
		\begin{minipage}[t]{0.52\textwidth}
			\begin{algorithm}[H]
				\caption{Training} \label{alg:training}
				\small
				\begin{algorithmic}[1]
					\Repeat
					\State Sample real data: $\mathbf{x}_1,  \mathbf{x}_2 \sim \mathcal{X}$
					\State Sample slice vectors: $\mathbf{v} \sim q(\mathbf{v}), \mathbf{w} \sim q(\mathbf{w})$
					\State $\delta = \mathcal{N}(\mathbf{w}^T\mathbf{x}_2 - \mathbf{w}^T\mathbf{x}_1; 0, \epsilon)$
					\State $\widebar{\mathcal{L}}_{\mathrm{grad}}^{\mathrm{slice}}(\bm{\phi};\epsilon) \approx (\mathbf{v}^{\top} \mathbf{g}_{\bm{\phi}}(\mathbf{x}_1))^2 + \delta (\mathbf{v}^{\top} \nabla_{\mathbf{x}_1} \mathbf{g}_{\bm{\phi}}(\mathbf{x}_1) \mathbf{v})$
					\State $\widebar{\mathcal{L}}_{\mathrm{id}}^{\mathrm{slice}}(\bm{\varphi};\epsilon) \approx s_{\bm{\varphi}}(\mathbf{x}_1)^2  - 2 \delta s_{\bm{\varphi}}(\mathbf{x}_1) $
					\State Update parameter $\bm{\phi}$ w.r.t. $- \nabla_{\bm{\phi}} \widebar{\mathcal{L}}_{\mathrm{grad}}^{\mathrm{slice}}(\bm{\phi};\epsilon)$ 
					\State Update parameter $\bm{\varphi}$ w.r.t. $- \nabla_{\bm{\varphi}} \widebar{\mathcal{L}}_{\mathrm{grad}}^{\mathrm{slice}}(\bm{\varphi};\epsilon)$ 
					\Until{converged}
				\end{algorithmic}
			\end{algorithm}
		\end{minipage}
		\hfill
		\begin{minipage}[t]{0.48\textwidth}
			\begin{algorithm}[H]
				\caption{Sampling} \label{alg:sampling}
				\small
				\begin{algorithmic}[1]
					\State Sample $\mathbf{x}(0)$ from any initial distribution
					\State Set sampling steps $T$ 
					\State Set constant step size $\eta$ 
					\For{$t=0, 1, \dotsc, T - 1$}
					\State $\mathbf{z} \sim \mathcal{N}(\mathbf{0}, \mathbf{I}_D)$
					\State $\Delta = \mathbf{g}_{\bm{\phi}}(\mathbf{x}(\eta t)) \eta  + \sqrt{s_{\bm{\varphi}}(\mathbf{x}(\eta t)) \eta } \mathbf{z}$
					\State $\mathbf{x}(\eta(t + 1)) = \mathbf{x}(\eta t) + \Delta$
					\EndFor
					\State \textbf{return} $\mathbf{x}(\epsilon T)$
				\end{algorithmic}
			\end{algorithm}
		\end{minipage}
	\end{figure}
	
	\section{Main Theory}
	\label{sec:theory}
	
	In this section, we provide theoretical support for Bellman Diffusion Dynamics, including steady-state and error analyses. We defer all proofs to Appendix~\ref{sec:main_proofs}.
	
	\subsection{Steady-State Analysis of Bellman Diffusion Dynamics}
	
	Let $p_t$ be the marginal density of Bellman Diffusion Dynamics given by Eq.~\eqref{eq:new sampling method}, starting from any initial density $p_0$. The following theorem shows that, regardless of the initial distribution $p_0$, $p_t$ converges to the stationary distribution, which is exactly $p_{\text{target}}(\rvx)$, as $t \rightarrow \infty$, at an exponential rate.
	\begin{theorem}[Convergence to the Steady State]\label{thm:stationary_dist}
		Let $p_{\mathrm{target}}$ be the target density satisfying Assumption~\ref{assumption_1}. Then, for any initial density $p_0$, we have the following KL and Wasserstein-2 bounds:
		\begin{align*}  W_2^2\big(p_t, p_{\mathrm{target}}\big) \lesssim \mathrm{KL}\big(p_t \Vert p_{\mathrm{target}}\big) \lesssim e^{-2\alpha t} \mathrm{KL}\big(p_0 \Vert p_{\mathrm{target}}\big). \end{align*}
		Here, $\alpha > 0$ is some constant determined by $p_{\mathrm{target}}$, and $\lesssim$ hides multiplicative constants that depend only on $p_{\mathrm{target}}$.
	\end{theorem}
	
	This theorem implies that as $t \to \infty$, $p_t \to p_{\mathrm{target}}$ in both KL and Wasserstein-2 senses. Thus, it justifies that by using our sampling method, which involves solving the SDE in Eq.~\eqref{eq:new sampling method}, we can ensure that samples will be obtained from the target distribution $p_{\mathrm{target}}$.
	
	\subsection{Error Analysis of Empirical Bellman Diffusion Dynamics}
	
	We let $p_{t;\bm{\phi},\bm{\varphi}}$ denote the marginal density from the empirical Bellman Diffusion Dynamics in Eq.~\eqref{eq:nn new sampling method}, starting from any initial density $p_0$. The following theorem extends the result in Theorem~\ref{thm:stationary_dist} by providing an error analysis. It accounts for network approximation errors in $\mathbf{g}_{\bm{\phi}}(\mathbf{x}) \approx \nabla p_{\mathrm{target}}(\mathbf{x})$ and $s_{\bm{\varphi}}(\mathbf{x}) \approx p_{\mathrm{target}}(\mathbf{x})$, and gives an upper bound on the Wasserstein-2 discrepancy between $p_{t;\bm{\phi},\bm{\varphi}}$ and $p_{\mathrm{target}}$.

	\begin{theorem}[Error Analysis of Neural Network Approximations]\label{thm:informal_error}
		Let $p_{\mathrm{target}}$ be the target distribution satisfying Assumptions~\ref{assumption_1} and \ref{assumption_2}. Suppose the dynamics in Eqs.~\eqref{eq:new sampling method} and \eqref{eq:nn new sampling method} start from the same initial  condition sampled from $p_0$. For any $\varepsilon > 0$, if $T=\mathcal O(\log 1/\varepsilon^2)$ and $\varepsilon_{\mathrm{est}}=\mathcal O\Big(\frac{\varepsilon}{\sqrt{T}e^{\frac{1}{2}LT}}\Big)$, such that
		$\norm{g_{\bm{\phi}}(\cdot) - \nabla p_{\mathrm{target}}(\cdot)}_\infty \leq \varepsilon_{\mathrm{est}}$ and $ \abs{s_{\bm{\varphi}}(\cdot) -  p_{\mathrm{target}}(\cdot)}_\infty  \leq \varepsilon_{\mathrm{est}}$,
		where $L > 0$ is the Lipschitz constant associated with $p_{\mathrm{target}}$, then
		\begin{align*}
			W_2(p_{T;\bm{\phi},\bm{\varphi}}, p_{\mathrm{target}})\leq \varepsilon.
		\end{align*}
	\end{theorem}
	From the above theorem, our dynamics can function as a standalone generative model, capable of learning the target distribution $p_{\mathrm{target}}$. Using advanced techniques such as \citet{chen2022sampling,de2022convergence,kim2023consistency,kim2024pagoda}, a tighter bound between $p_{T;\bm{\phi},\bm{\varphi}}$ and $p_{\mathrm{target}}$ in $W_2$ or other divergences could be achieved. Moreover, discrete-time versions of both Theorems~\ref{thm:stationary_dist} and \ref{thm:informal_error} can be derived with more advanced analysis. However, we defer this to future work, as the current focus is on establishing the core principles.
	
	\section{Related Work}
	
	\paragraph{Deep generative modelings.} Bellman Diffusion stands as a new class of generative models. Sec.~\ref{sec:current dgms} revisits modern DGMs and re-examines whether their modeling satisfies the linear property. Additional details on related works concerning DGMs can be found in Appendix~\ref{sec:dgm_review}.
	
	\paragraph{Markov decision processes.} Limited by the linearity of the distributional Bellman equation, Previous works~\citep{bellemare2017distributional,hessel2018rainbow,dabney2018implicit} in planning and distributional RL have relied on conventional generative models to represent state-level return distributions. For instance, the widely used C51~\citep{bellemare2017distributional} is a histogram model, resulting in discrete approximation errors. In contrast, Bellman Diffusion is a new type of diffusion model that serves as an expressive distribution approximator without discretization errors. As shown in Sec.~\ref{sec:experiments}, our experiments demonstrate that Bellman Diffusion achieves significantly faster and much more stable convergence properties.

	\section{Experiments}
	\label{sec:experiments}
	
	\begin{figure*}
		\centering
		\includegraphics[width=0.8\textwidth, height=0.12\textheight]{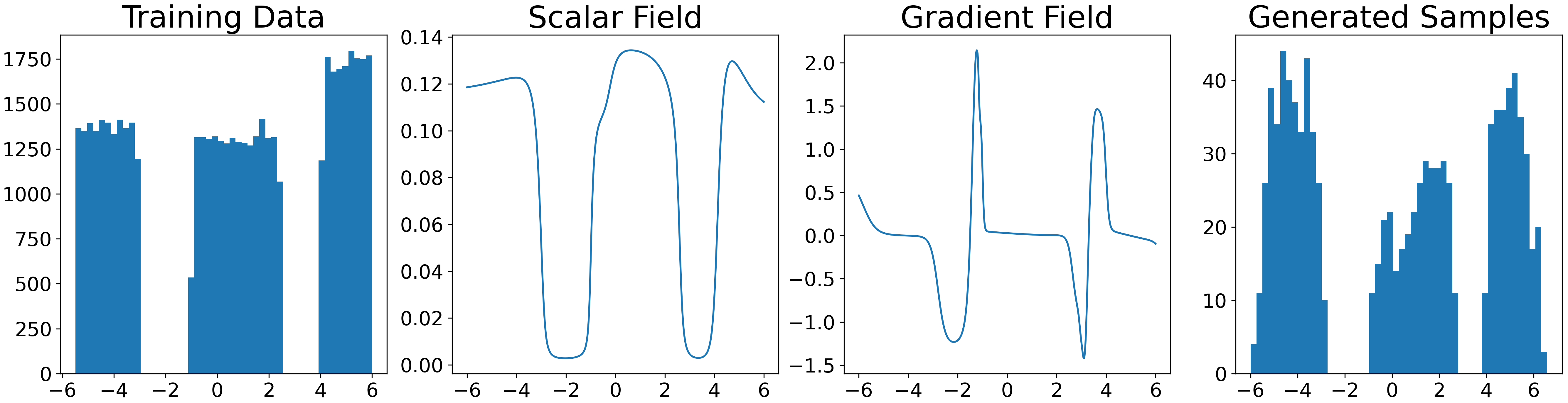}
		\caption{\textit{Bellman Diffusion captures the uniform distribution supported on disjoint spans}. The leftmost subfigure presents the training data histogram, while the next three show the estimated density, derivative functions, and samples generated by Bellman Diffusion.}
		\label{fig:toy left results}
	\end{figure*}
	
	\begin{figure*}
		\centering
		\includegraphics[width=0.8\textwidth, height=0.12\textheight]{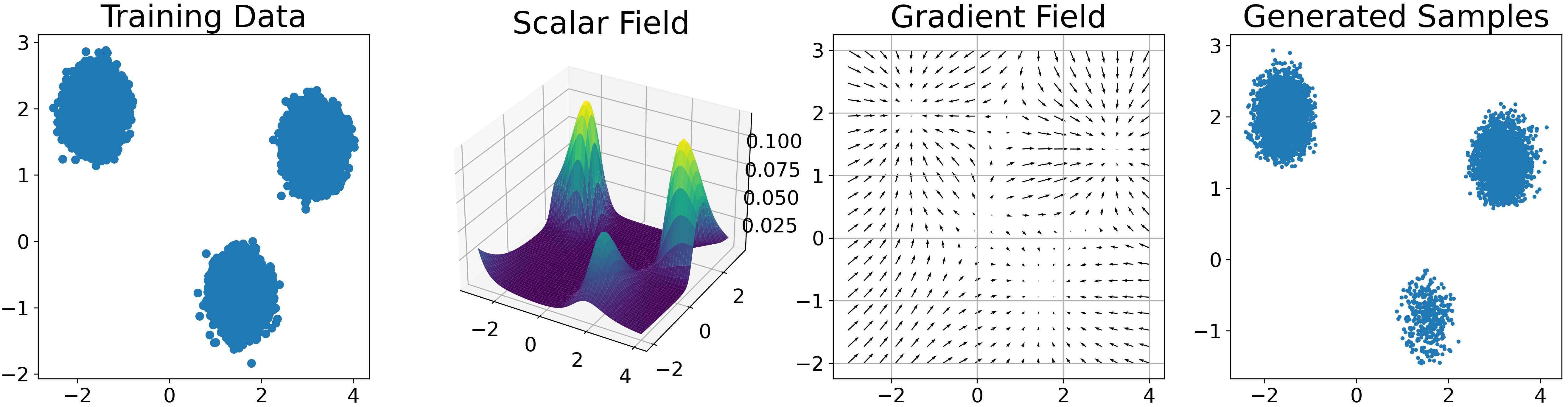}
		\caption{\textit{Bellman Diffusion learns the unbalanced Gaussian mixture}, which is hard for score-based models. The subfigures, from left to right, display the training data, estimated scalar and gradient fields, and samples generated by our Bellman Diffusion.}
		\label{fig:toy middle results}
	\end{figure*}
	
	To verify the effectiveness of our method: Bellman Diffusion, we have conducted extensive experiments on multiple synthetic and real benchmarks across different tasks (e.g., generative modeling and MDPs). The aim of our experiments is to verify that Bellman Diffusion is both a capable generative model and an effective distributional RL model.  We also place the experiment setup in Appendix~\ref{appendix:experiment setup} and other minor experiments in Appendix~\ref{appendix:extra experiments}.
	
	\subsection{Synthetic Datasets}\label{sec:experiments-toy}
	
	In this part, we aim to show that Bellman Diffusion can accurately estimate the scalar and gradient fields $\nabla p_{\mathrm{target}}(\rvx), p_{\mathrm{target}}(\rvx)$ and the associated sampling dynamics can recover the data distribution in terms of the estimation models $\mathbf{g}_{\bm{\phi}}(\mathbf{x}), s_{\bm{\varphi}}(\mathbf{x})$. For visualization purpose, we will adopt low-dimensional synthetic data (i.e., $D = 1$, or $2$) in the studies. 
	\paragraph{1-dimensional uniform distribution.} As shown in the leftmost subfigure of Fig.~\ref{fig:toy left results}, we apply Bellman Diffusion to a uniform distribution over three disjoint spans. A key challenge is approximating the discontinuous data distribution using continuous neural networks $ g_{\bm{\phi}}(\rvx) $ and $ s_{\bm{\varphi}}(\rvx) $.
	
	Interestingly, the results in Fig.~\ref{fig:toy left results} indicate that the estimated field models $ \mathbf{g}_{\bm{\phi}}(\mathbf{x}) $ and $ s_{\bm{\varphi}}(\mathbf{x}) $ closely approximate the correct values on the support (e.g., $ [-1.0, 2.0] $) and behave reasonably in undefined regions. For instance, $ g_{\bm{\phi}}(\rvx) $ resembles a negative sine curve on $ [-4.5, -0.5] $, aligning with the definitions of one-sided derivative. Notably, our Bellman Diffusion Dynamics yield a generation distribution nearly matching the training samples, demonstrating effectiveness in learning discontinuous data distributions.
	
	\paragraph{2-Dimensional Mixture of Gaussian (MoG).} Bellman Diffusion effectively approximates the density and gradient fields for multimodal distributions, even with unbalanced weights, which are not recognizable for SGMs.  We investigate this using a MoG distribution of three modes with differing weights ($0.45$, $0.45$, $0.1$), as shown in the leftmost subfigure of Fig.~\ref{fig:toy middle results}.
	
	\begin{figure*}
		\centering
		\includegraphics[width=0.8\textwidth, height=0.12\textheight]{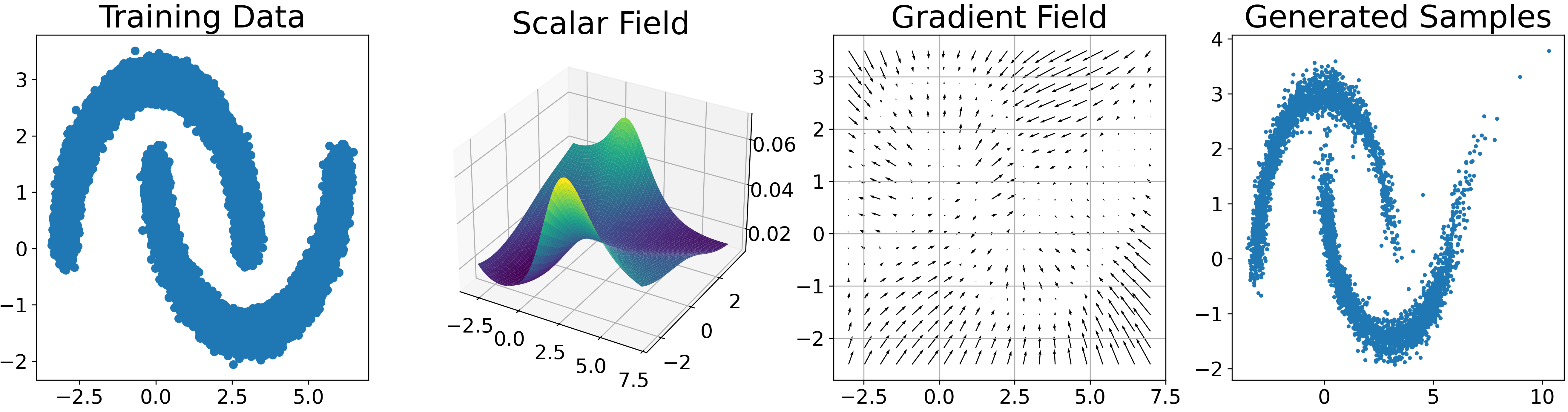}
		\caption{\textit{Bellman Diffusion learns unusually clustered data}. The subfigures, from left to right, show the training data, estimated density field, gradient field, and generated samples.}
		\label{fig:toy right results}
	\end{figure*}
	
	The right three subfigures of Fig.~\ref{fig:toy middle results} present the experimental results, demonstrating accurate estimation of both the scalar and gradient fields for the target data distribution $ p_{\mathrm{target}}(\rvx) $ and $ \nabla p_{\mathrm{target}}(\rvx) $. The three clustering centers of the training data correspond to the three density peaks in the scalar field (leftmost subfigure) and the critical points in the gradient field (middle subfigure). Notably, Bellman Diffusion successfully recovers the unbalanced modes of the target distribution and accurately estimates the density and gradient fields, even in low-density regions—a challenge for SGMs~\citep{song2019generative} due to its score design.
	
	\paragraph{2-dimensional moon-shaped data.} To demonstrate the ability of Bellman Diffusion to learn distributions with disjoint supports, we test it on the two moon dataset, where samples cluster into two disjoint half-cycles, as shown in the leftmost subfigure of Fig.~\ref{fig:toy right results}.
	
	The right three parts of Fig.~\ref{fig:toy right results} shows that the estimated scalar and gradient fields $ p_{\mathrm{target}}(\rvx) $ and $ \nabla p_{\mathrm{target}}(\rvx) $ match the training samples, with correctly positioned density peaks (leftmost subfigure) and critical points (middle subfigure). Our diffusion sampling dynamics accurately recover the shape of the training data, even in low-density regions. Thus, we conclude that Bellman Diffusion is effective in learning from complex data.
	
	\subsection{High-dimensional Data Generation}
	In this section, we follow the common practice~\citep{song2020sliced} to examine the scalability of our approach across multiple UCI tabular datasets~\citep{asuncion2007uci}, including Abalone, Telemonitoring, Mushroom, Parkinson's, and Red Wine. We apply several preprocessing steps to these datasets, such as imputation and feature selection, resulting in data dimensions of $7$, $16$, $5$, $15$, and $10$, respectively. For evaluation metrics, we utilize the commonly used Wasserstein distance~\citep{ruschendorf1985wasserstein} and maximum mean discrepancy (MMD)~\citep{dziugaite2015training}. The performance of a generative model is considered better when both metrics are lower.
	
	The experimental results are presented in Table~\ref{tab:tabular comparison}. We observe that, regardless of the dataset or metric, Bellman Diffusion performs competitively with DDPM~\citep{ho2020denoising,song2019generative}, a diffusion model known for its scalability. For instance, although our model falls short of DSM by just $0.139$ points on the Telemonitoring dataset (as measured by the Wasserstein distance), it achieves an even better MMD score on the Abalone dataset. 
	
	We further demonstrate in Appendix~\ref{sec:img_gen} that Bellman Diffusion is compatible with VAE~\citep{kingma2013auto}, allowing latent generative model training similar to latent diffusion models~\citep{rombach2022high} for higher-resolution image generation, highlighting the scalability of Bellman Diffusion. These results confirm that Bellman Diffusion is a scalable generative model.
	
	\begin{table}[th]
		\centering
		\small
		\scalebox{0.9}{  \begin{tabular}{c|cc|cc}
				\hline
				\multirow{2}{*}{Dataset} & \multicolumn{2}{c|}{Denoising Diffusion Models} & \multicolumn{2}{c}{Our Model: Bellman Diffusion} \\ \cline{2-5}
				& Wasserstein $\downarrow$& MMD ($10^{-3}$) $\downarrow$& Wasserstein $\downarrow$& MMD ($10^{-3}$)$\downarrow$\\ \hline
				Abalone & $0.975$ & $5.72$ & $0.763$ & $5.15$ \\
				Telemonitoring & $2.167$ & $10.15$ & $2.061$ & $9.76$ \\ 
				Mushroom & $1.732$ & $4.29$ & $1.871$ & $5.12$ \\
				Parkinsons & $0.862$ & $3.51$  & $0.995$ & $3.46$ \\ 
				Red Wine & $1.151$ & $3.83$ & $1.096$ & $3.91$ \\ \hline
			\end{tabular}
		}
		\caption{Performances on multiple high-dimensional datasets, showing that Bellman Diffusion is a capable generative model in high dimensions.}
		\label{tab:tabular comparison}
	\end{table}

	\subsection{Applications to RL}
	
	\begin{figure*}
		\centering
		\begin{tabular}{ccc}
			\begin{subfigure}[b]{0.32\textwidth}
				\includegraphics[width=\textwidth]{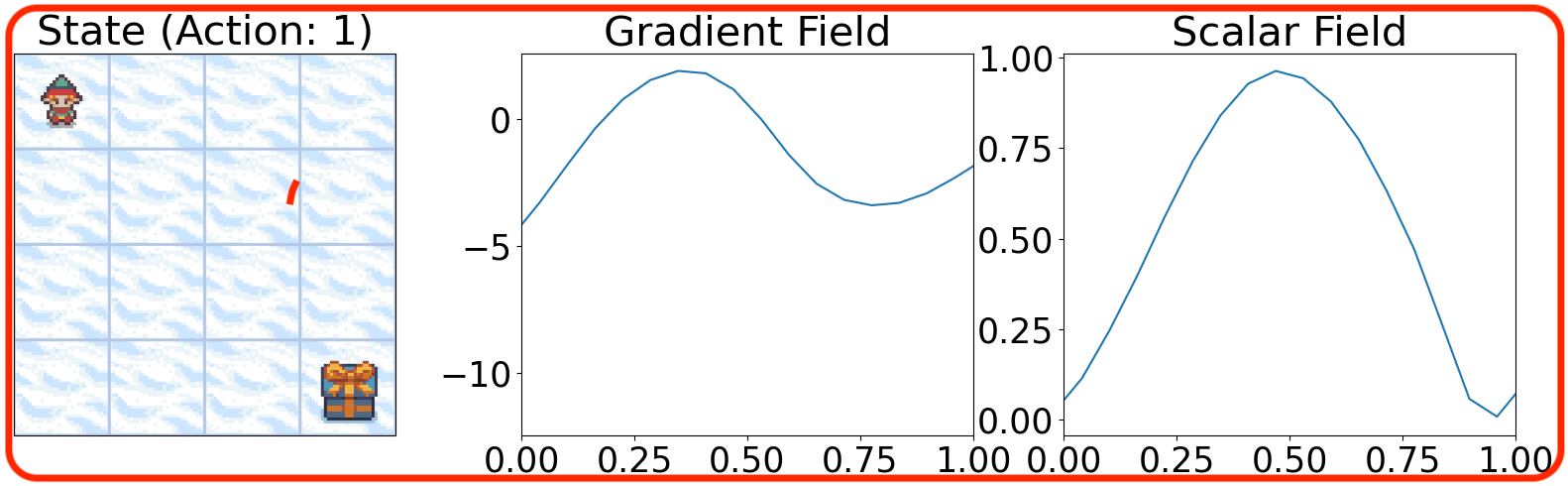}
			\end{subfigure}
			\begin{subfigure}[b]{0.32\textwidth}
				\includegraphics[width=\textwidth]{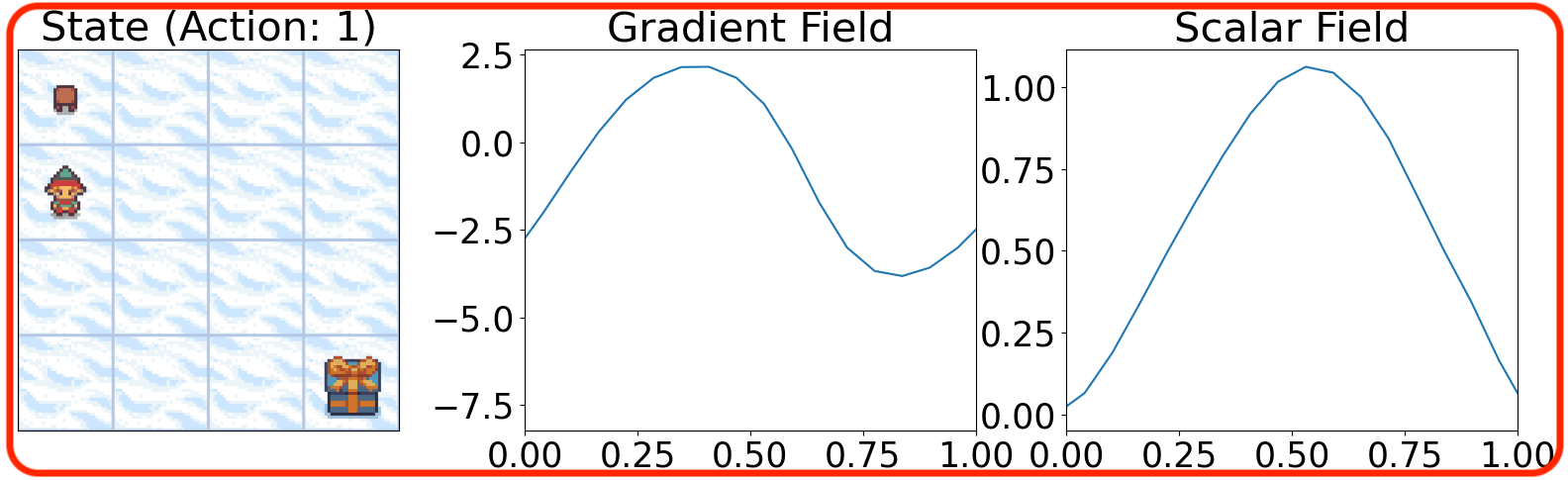}
			\end{subfigure}
			\begin{subfigure}[b]{0.32\textwidth}
				\includegraphics[width=\textwidth]{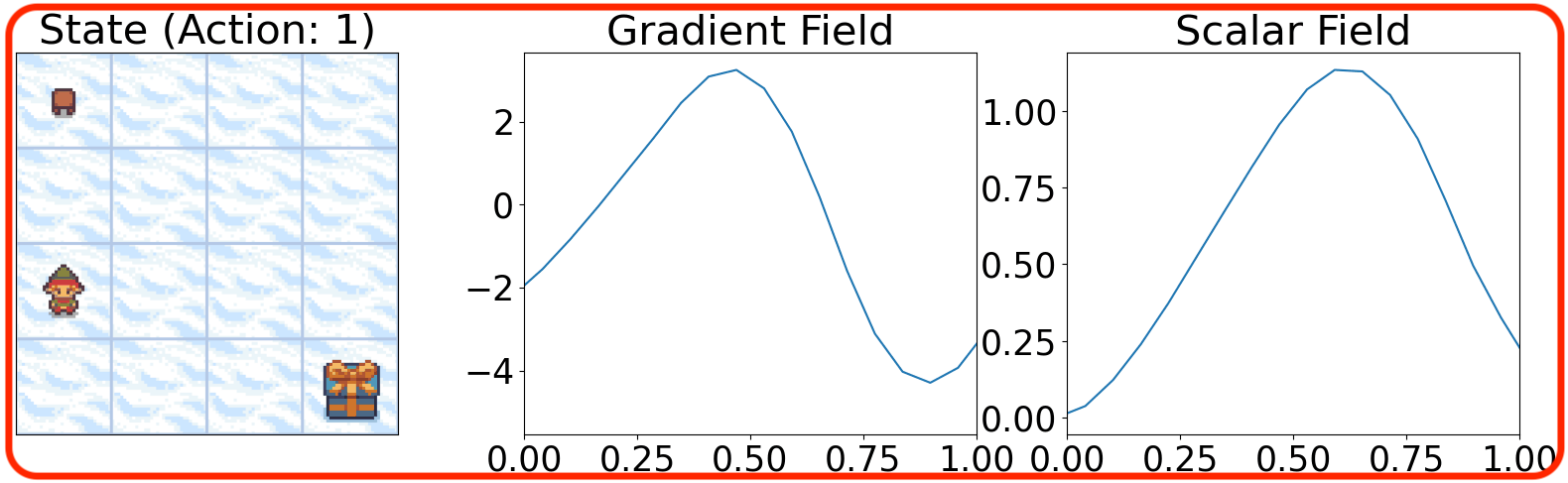}
			\end{subfigure} \\
			
			\begin{subfigure}[b]{0.32\textwidth}
				\centering
				\includegraphics[width=\textwidth]{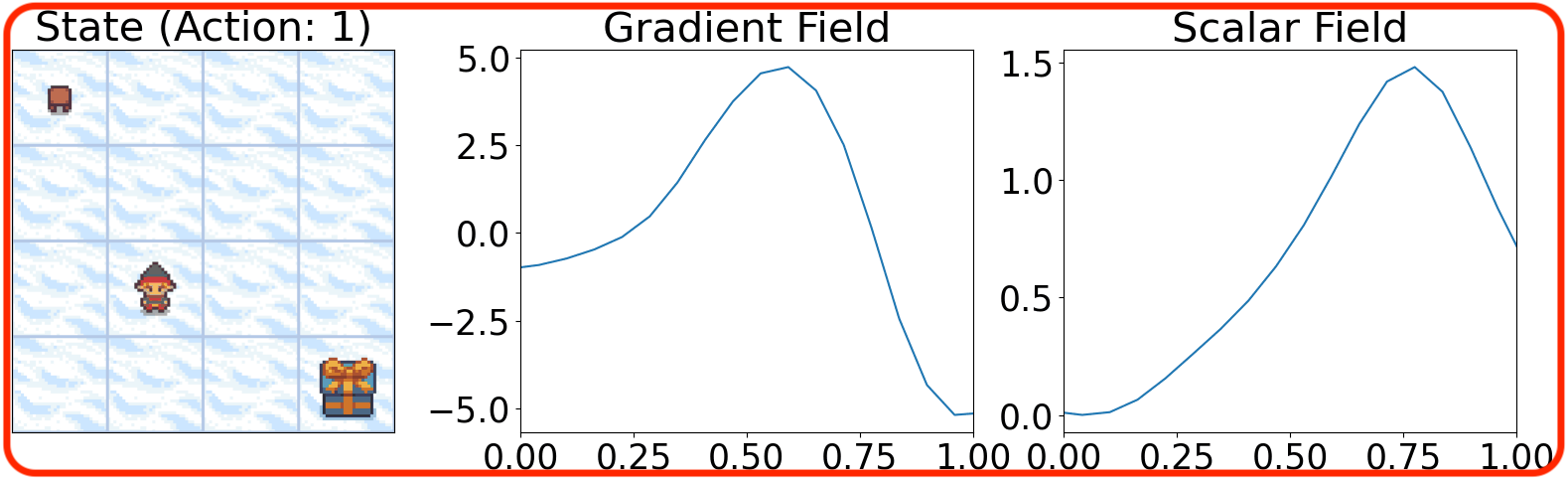}
			\end{subfigure}
			\begin{subfigure}[b]{0.32\textwidth}
				\centering
				\includegraphics[width=\textwidth]{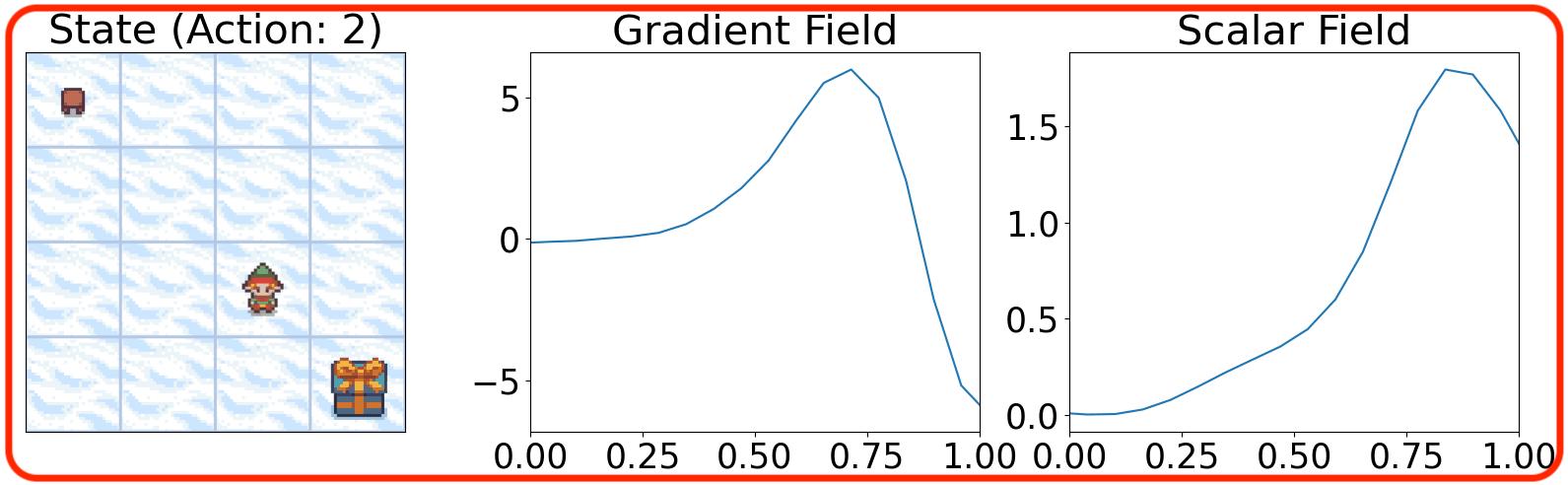}
			\end{subfigure}
			\begin{subfigure}[b]{0.32\textwidth}
				\centering
				\includegraphics[width=\textwidth]{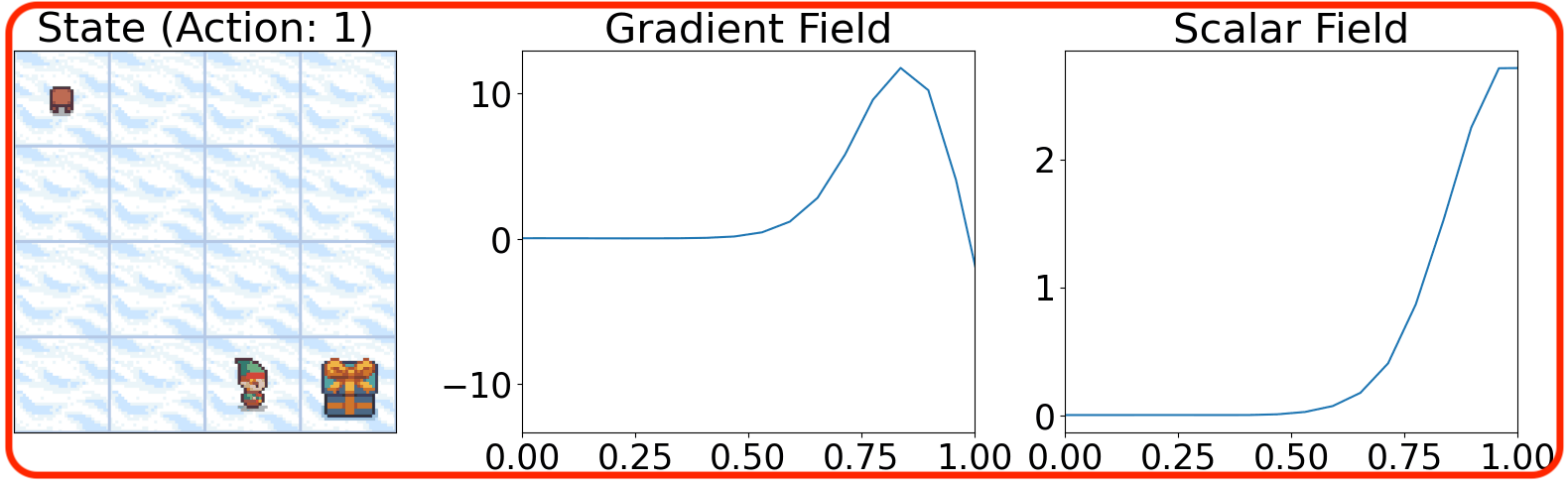}
			\end{subfigure}
		\end{tabular}
		\caption{The $2 \times 3$ subfigures, arranged from left to right and top to bottom, show a full trajectory of 
			Bellman Diffusion, interacting with a maze environment. Each subfigure consists of the state on the left, gradient field in the middle, and scalar field on the right.}
		\label{fig:maze experiment}
	\end{figure*}
	
	\begin{figure*}
		\centering
		\begin{subfigure}[b]{0.49\textwidth}
			\centering
			\includegraphics[width=\textwidth]{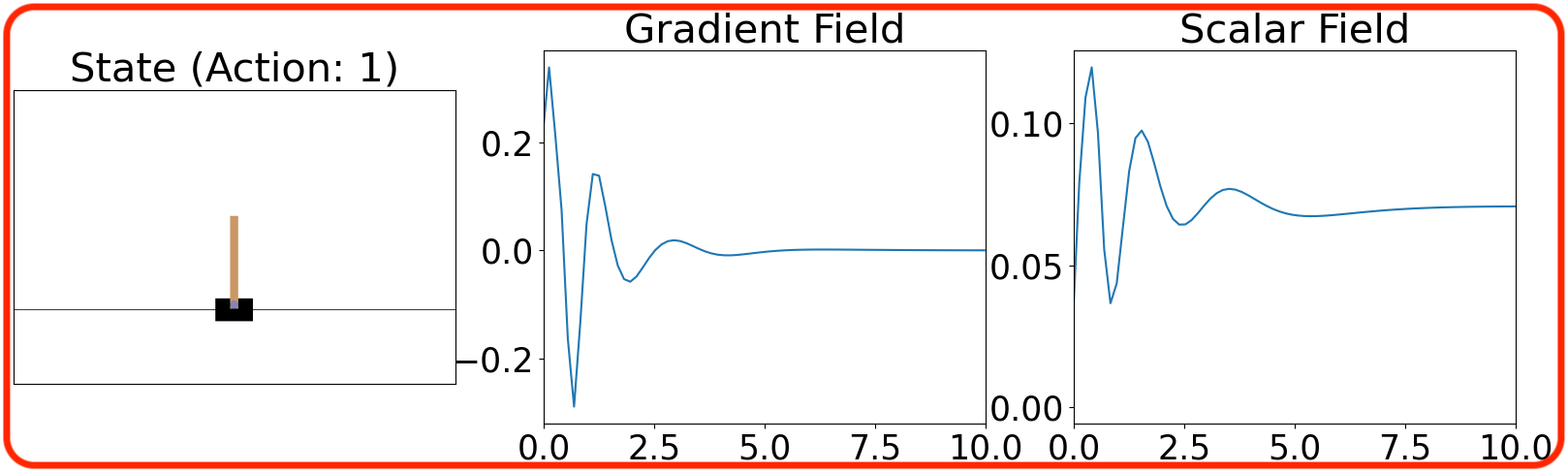}
		\end{subfigure}
		\begin{subfigure}[b]{0.49\textwidth}
			\centering
			\includegraphics[width=\textwidth]{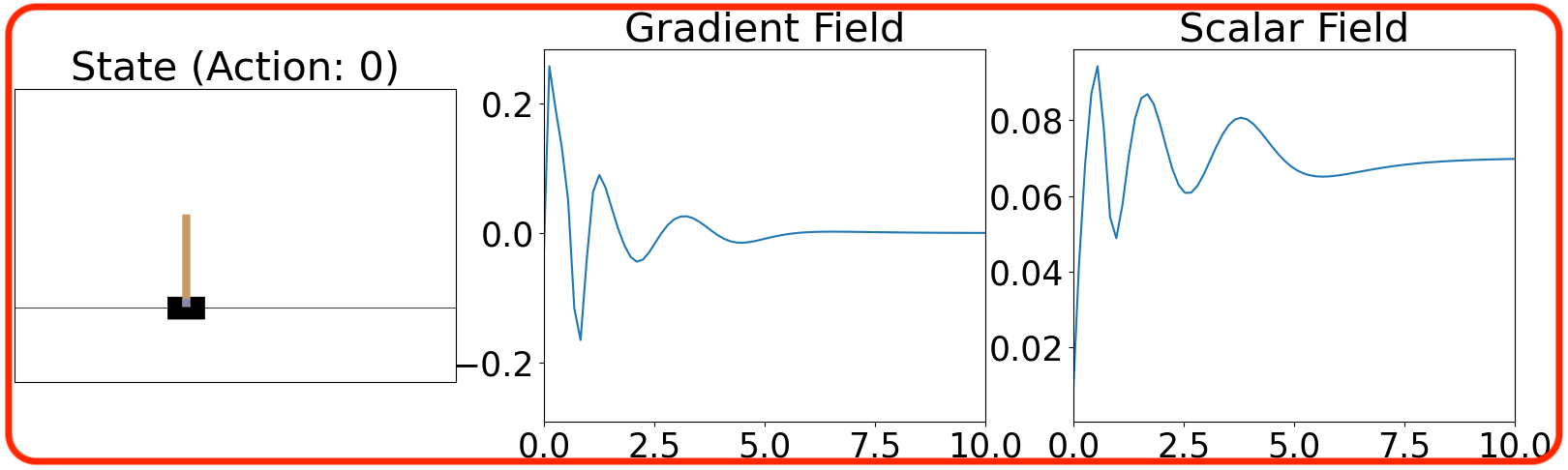}
		\end{subfigure}
		\caption{The left and right subfigures respectively show the initial and terminal states of Bellman Diffusion, interacting with an environment of balance control. Every subfigure is composed of the observation on the left, gradient field in the middle, and scalar field on the right.}
		\label{fig:cartpole experiment}
	\end{figure*}
	We demonstrate that Bellman Diffusion is effective for distributional RL, a classical task in MDPs. A method successful in RL can also handle simpler tasks (e.g., planning). Compared to previous baselines (e.g., C51~\citep{bellemare2017distributional}), our method learns a continuous return distribution at each state, avoiding errors from discrete approximations. We expect better convergence properties than the baseline, as our approach minimizes error propagation along state transitions.
	
	\textbf{Case studies.}
	We apply Bellman Diffusion to two OpenAI Gym environments~\citep{brockman2016openai}: Frozen Lake and Cart Pole. Concrete implementations are detailed in Appendix~\ref{sec:apply to mdp}. Frozen Lake is a maze where actions (e.g., moving up) may yield unexpected outcomes (e.g., moving left), while Cart Pole involves balancing a pole on a movable car. 
	\begin{wrapfigure}{r}{0.45\textwidth}
		\centering
		\vspace{-0.3cm}
		\includegraphics[width=0.45\textwidth]{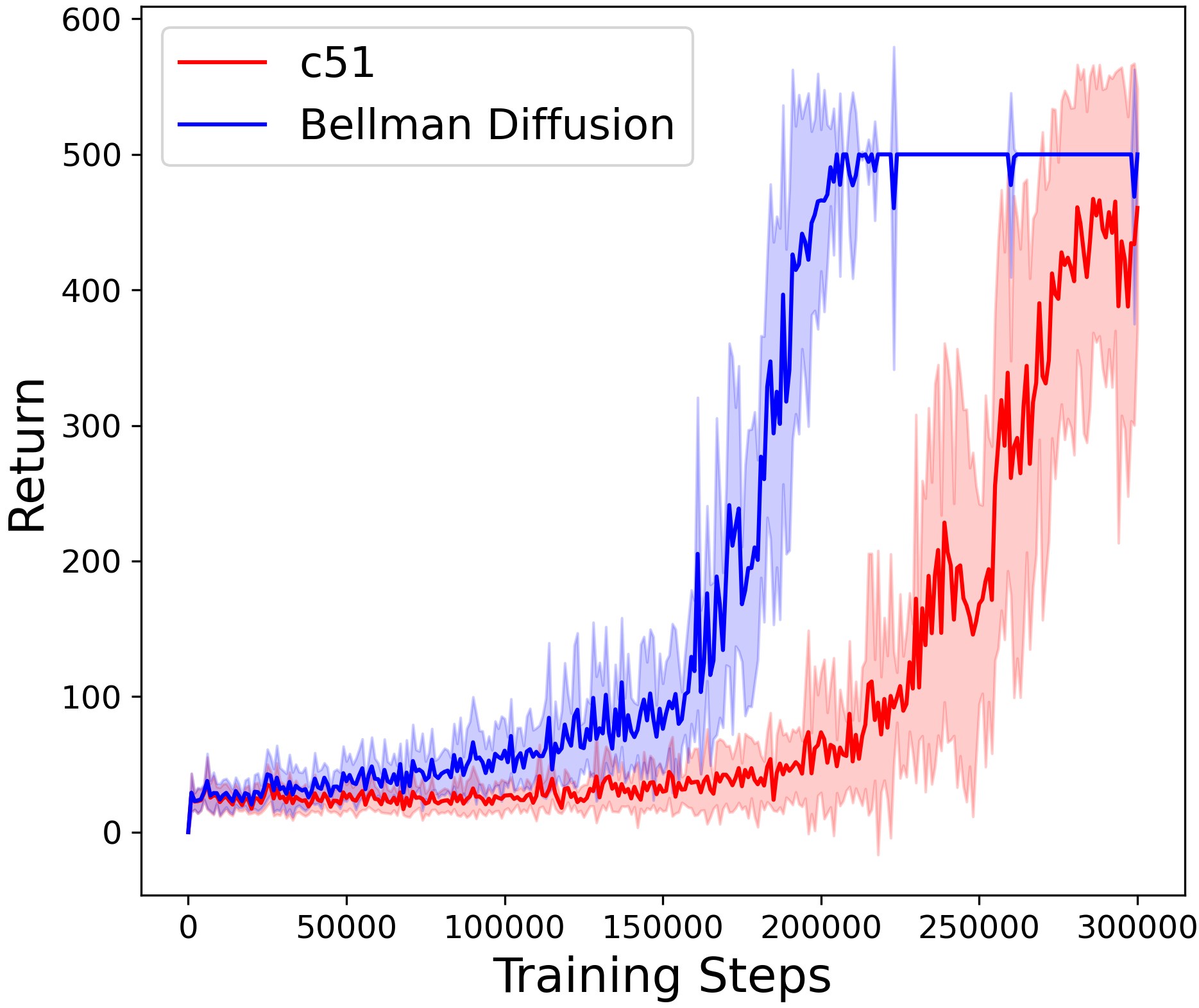}
		\caption{The return dynamics on the Cart Pole, with the colored areas as the confidence interval and the maximum return as $500$.}
		\label{fig:convergence_results}
		\vspace{-1.0cm}
	\end{wrapfigure} 
	Results in Figs.~\ref{fig:maze experiment} and \ref{fig:cartpole experiment} show that Bellman Diffusion accurately estimates state-level return distributions and their derivatives. For instance, as the agent approaches the goal in the maze, the expected return shifts from $0.5$ to $1$, reflecting that the agent receives no rewards until it reaches the goal.
	
	\textbf{Faster and more stable convergence.} With the same model sizes and different random seeds, Bellman Diffusion and C51, a widely used distributional RL method, are both run on the environment of Cart Pole $10$ times. The results of return dynamics over training steps are shown in Fig.~\ref{fig:convergence_results}. We can see that, while both models have the potential to reach the maximum return in the end, Bellman Diffusion converges much faster than C51, with highly stable dynamics. These results confirm our previous conjecture about the superiority of Bellman Diffusion.
	
	\section{Conclusion}
	
	In this work, we highlight the limitations of modern DGMs, including emerging SGMs, in the context of MDPs and distributional RL. By identifying the critical necessity for linearity in these applications, we introduce Bellman Diffusion, a novel DGM that effectively maintains this linearity by directly modeling gradient and scalar fields. Through the development of divergence measures, we train neural network proxies for these fields, facilitating a new sampling dynamic: Bellman Diffusion Dynamics, via a SDE that ensures convergence to the target distribution. Our experiment results show that Bellman Diffusion not only achieves accurate field estimations and generates high-quality images, but also shows improved performance in classical distributional RL tasks. These findings suggest that Bellman Diffusion is a promising approach for advancing the integration of DGMs within the RL frameworks.
	
	\bibliography{iclr2025_conference}
	\bibliographystyle{iclr2025_conference}
	
	\newpage
	\appendix
	
	\begingroup
	\hypersetup{colorlinks=false, linkcolor=red}
	\hypersetup{pdfborder={0 0 0}}
	\renewcommand\ptctitle{}
	
	\part{Appendix} 
	\parttoc 
	
	\endgroup
	
	\newpage
	\section{More Related Works on DGMs}\label{sec:dgm_review}
	Deep generative models (DGMs) have gained significant attention in recent years due to their ability to learn complex data distributions and generate high-fidelity samples. This literature review covers several prominent categories of DGMs, including Variational Autoencoders (VAEs), Generative Adversarial Networks (GANs), energy-based methods, flow-based methods, and diffusion models.
	
	\paragraph{Variational Autoencoders (VAEs).}
	Variational Autoencoders (VAEs) are a class of generative models that leverage variational inference to approximate the posterior distribution of latent variables given the data. The VAE framework is based on the evidence lower bound (ELBO), which can be expressed as:
	\begin{equation*}
		\mathcal{L}(\bm{\theta}, \bm{\phi}; \rvx) = \mathbb{E}_{q_{\bm{\phi}}(\rvz|\rvx)}[\log p_{\bm{\theta}}(\rvx|\rvz)] - \mathcal D_{\mathrm{KL}}(q_{\bm{\phi}}(\rvz|\rvx) || p(\rvz)),
	\end{equation*}
	where $ q_{\bm{\phi}}(\rvz|\rvx) $ is the approximate posterior, $ p_{\bm{\theta}}(\rvx|\rvz) $ is the likelihood, and $\mathcal D_{\mathrm{KL}} $ denotes the Kullback-Leibler divergence. VAEs have shown remarkable success in generating images and other complex data types \cite{kingma2013auto}.
	
	\paragraph{Generative Adversarial Networks (GANs).}
	Generative Adversarial Networks (GANs) consist of two neural networks, a generator $ G $ and a discriminator $ D $, that compete against each other. The generator aims to create realistic samples $ G(\rvz) $ from random noise $ \rvz $, while the discriminator attempts to distinguish between real samples $ \rvx $ and generated samples $ G(\rvz) $. The objective function for GANs can be formulated as:
	
	\begin{equation*}
		\min_G \max_D \mathbb{E}_{\rvx \sim p_{\mathrm{target}}(\rvx)}[\log D(\rvx)] + \mathbb{E}_{\rvz \sim p_{\rvz}(\rvz)}[\log(1 - D(G(\rvz)))],
	\end{equation*}
	
	where $ p_{\mathrm{data}}(\rvx) $ is the data distribution and $ p_{\rvz}(\rvz) $ is the prior distribution on the noise. GANs have become popular for their ability to produce high-quality images and have been applied in various domains~\cite{goodfellow2020generative}.
	
	\paragraph{Energy-Based Models}
	Energy-based models (EBMs) define a probability distribution through an energy function $ E(\rvx) $ that assigns lower energy to more probable data points. The probability of a data point is given by:
	\begin{equation*}
		p(\rvx) = \frac{1}{Z} \exp(-E(\rvx)),
	\end{equation*}
	where $ Z = \int \exp(-E(\rvx)) d\rvx $ is the partition function. Training EBMs typically involves minimizing the negative log-likelihood of the data~\citep{lecun2006tutorial}. They have been successfully applied in generative tasks, including image generation and modeling complex data distributions.
	
	\paragraph{Flow-Based Methods}
	Flow-based methods, such as Normalizing Flows (NFs), learn a bijective mapping between a simple distribution $ \rvz $ and a complex data distribution $ \rvx $ through a series of invertible transformations. The probability density of the data can be expressed as:
	\begin{equation*}
		p(\rvx) = p(\rvz) \left| \det \frac{\partial \rvf^{-1}}{\partial \rvx} \right|,
	\end{equation*}
	where $ \rvf $ is the invertible transformation from $ \rvz $ to $ \rvx $. Flow-based models allow for efficient exact likelihood estimation and have shown promise in generating high-quality samples \citep{rezende2015variational}.
	
	\paragraph{Diffusion Models}
	Diffusion models are a class of generative models that learn to generate data by reversing a gradual noising process. The generative process can be described using a stochastic differential equation (SDE):
	\begin{equation*}
		\diff\rvx_t = \rvf(\rvx_t, t) \diff d t + g(t) \diff d\rvw_t,
	\end{equation*}
	where $ \rvw_t $ is a Wiener process, and $ \rvf(\rvx_t, t) $ and $ g(t) $ are functions defining the drift and diffusion terms, respectively. The model learns to recover the data distribution from noise by training on the denoising score matching objective~\citep{ho2020denoising,song2021scorebased}. Diffusion models have recently gained attention for their impressive image synthesis capabilities.

	\section{Theoretical Results and Proofs for Sec.~\ref{sec:method overview}}\label{sec:proofs}
	\subsection{Motivation of the Proposed Dynamics 
		in Eq.~\eqref{eq:new sampling method}}\label{sec:motivation}
	\paragraph{$1$-dimensional Case.}
	Let us first consider the one-dimensional case:
	\begin{equation*}
		\diff x(t) = f(x(t)) \diff t + g(x(t)) \diff w(t).
	\end{equation*}
	Based on the Fokker–Planck equation~\cite{risken1996fokker}, the probability distribution $p(x, t)$ of dynamics $x(t)$ satisfies
	\begin{align*}
		\frac{\partial p(x, t)}{\partial t} & = - \frac{\partial}{\partial x} \Big( f(x) p(x, t)  \Big) + \frac{1}{2} \frac{\partial^2}{\partial x^2} \Big( g^2(x) p(x, t)  \Big) \nonumber \\
		& = \frac{\partial}{\partial x}\Big( -f(x) p(x, t) + \frac{1}{2} \frac{\partial }{\partial x}\Big( g^2(x) p(x, t) \Big) \Big).
	\end{align*}
	
	Suppose that the density $p(x, t)$ converges as $t \rightarrow \infty$, then we have $	\partial p(x, t) / \partial t \mid_{t \rightarrow \infty} = 0$. As a result, the above equality indicates that
	\begin{equation*}
		-f(x) p(x, \infty) + \frac{1}{2} \frac{\diff}{\diff x}\Big( g^2(x) p(x, \infty) \Big) = C.
	\end{equation*}
	Suppose the constant $C$ is $0$, then we have
	\begin{equation*}
		g^2(x) \frac{\diff p(x, \infty)}{\diff x} = \Big(2f(x) - \frac{\diff g^2(x)}{\diff x} \Big) p(x, \infty).
	\end{equation*}
	One way to make this equality hold is to have the below setup:
	\begin{equation*}
		\left\{\begin{aligned}
			g^2(x) & = p(x, \infty) \\
			f(x) & = \frac{1}{2}\Big(  \frac{\diff p(x, \infty)}{\diff x} + \frac{\diff g^2(x)}{\diff x} \Big)  \Big) =  \frac{\diff p(x, \infty)}{\diff x}.
		\end{aligned}\right.
	\end{equation*}
	Therefore, the following dynamics:
	\begin{equation*}
		\diff x(t) =  \frac{\diff p_{\mathrm{target}} (x(t))}{\diff x} \diff t + \sqrt{p_{\mathrm{target}}(x(t))} \diff w(t),
	\end{equation*}
	will converge to distribution $p_{\mathrm{target}}(x)$ as $t \rightarrow \infty$, regardless of the initial distribution $p(x, 0)$.
	
	\paragraph{$D$-dimensional Case.}
	For the general situation, the dynamics will be
	\begin{equation*}
		\diff\mathbf{x}(t) = \nabla_{\mathbf{x}(t)} p_{\mathrm{target}} (\rvx(t)) \diff t +  \sqrt{p_{\mathrm{target}} (\rvx(t))} \diff\bm{\omega}(t), 
	\end{equation*}
	Let us check this expression. Firstly, the Fokker–Planck equation indicates that
	\begin{equation*}
		\begin{aligned}
			\frac{\partial p(\mathbf{x}, t)}{\partial t} & = -\nabla_{\mathbf{x}} \cdot \big( p(\mathbf{x}, t) \nabla_{\mathbf{x}} p_{\mathrm{target}}(\rvx)  \big) + \frac{1}{2} \Delta_{\mathbf{x}} \big( p_{\mathrm{target}}(\rvx) p(\mathbf{x}, t) \big) \\
			& = \nabla_{\mathbf{x}} \cdot \Big( - p(\mathbf{x}, t) \nabla_{\mathbf{x}} p_{\mathrm{target}}(\rvx)  + \frac{1}{2} \nabla_{\mathbf{x}} \big( p_{\mathrm{target}}(\rvx) p(\mathbf{x}, t) \big) \Big),
		\end{aligned}
	\end{equation*}
	where $\Delta_{\mathbf{x}} = \nabla_{\mathbf{x}} \cdot \nabla_{\mathbf{x}} $ is the Laplace operator. With the Leibniz rule, we have
	\begin{equation*}
		\begin{aligned}
			& - p(\mathbf{x}, t) \nabla_{\mathbf{x}} p_{\mathrm{target}}(\rvx)  + \frac{1}{2} \nabla_{\mathbf{x}} \big( p_{\mathrm{target}}(\rvx) p(\mathbf{x}, t) \big) \\
			=& - p(\mathbf{x}, t) \nabla_{\mathbf{x}} p_{\mathrm{target}}(\rvx) + \frac{1}{2} p(\mathbf{x}, t) \nabla_{\mathbf{x}} p_{\mathrm{target}}(\rvx) + \frac{1}{2} p_{\mathrm{target}}(\rvx) \nabla_{\mathbf{x}} p(\mathbf{x}, t) \\
			=& \frac{1}{2} p_{\mathrm{target}}(\rvx) \nabla_{\mathbf{x}} p(\mathbf{x}, t)  - \frac{1}{2} p(\mathbf{x}, t) \nabla_{\mathbf{x}} p_{\mathrm{target}}(\rvx).
		\end{aligned}
	\end{equation*}
	Combining the above two equations, we get
	\begin{equation*}
		\frac{\partial p(\mathbf{x}, t)}{\partial t} = \frac{1}{2} \nabla_{\mathbf{x}} \cdot \Big( p_{\mathrm{target}}(\rvx) \nabla_{\mathbf{x}} p(\mathbf{x}, t) - p(\mathbf{x}, t) \nabla_{\mathbf{x}} p_{\mathrm{target}}(\rvx) \Big).
	\end{equation*}
	By applying the Leibniz rule to divergence operators, we have
	\begin{equation*}
		\left\{\begin{aligned}
			\nabla_{\mathbf{x}} \cdot \big( p_{\mathrm{target}}(\rvx) \nabla_{\mathbf{x}} p(\mathbf{x}, t) \big) & = \big< \nabla_{\mathbf{x}} p_{\mathrm{target}}(\rvx), \nabla_{\mathbf{x}} p(\mathbf{x}, t) \big> + p_{\mathrm{target}}(\rvx)  \Delta_{\mathbf{x}} p(\mathbf{x}, t) \\
			\nabla_{\mathbf{x}} \cdot \big( p(\mathbf{x}, t) \nabla_{\mathbf{x}} p_{\mathrm{target}}(\rvx)  \big) & = \big< 	\nabla_{\mathbf{x}} p(\mathbf{x}, t), \nabla_{\mathbf{x}} p_{\mathrm{target}}(\rvx)  \big> + p(\mathbf{x}, t)  \Delta_{\mathbf{x}} p_{\mathrm{target}}(\rvx)
		\end{aligned}\right..
	\end{equation*}
	Therefore, the original partial differential equation (PDE) can be simplified as
	\begin{equation*}
		\frac{\partial p(\mathbf{x}, t)}{\partial t} = \frac{1}{2}\Big( p_{\mathrm{target}}(\rvx)  \Delta_{\mathbf{x}} p(\mathbf{x}, t) - p(\mathbf{x}, t)  \Delta_{\mathbf{x}} p_{\mathrm{target}}(\rvx) \Big).
	\end{equation*}
	Since the dynamics will converge, we set $p_{\mathrm{target}}(\rvx) = p(\mathbf{x}, \infty)$. Then, we get
	\begin{equation*}
		\frac{\partial p(\mathbf{x}, t)}{\partial t} \Big|_{t \rightarrow \infty} =  \frac{1}{2}\Big( p(\mathbf{x}, \infty)  \Delta_{\mathbf{x}} p(\mathbf{x}, \infty) - p(\mathbf{x}, \infty)  \Delta_{\mathbf{x}} p(\mathbf{x}, \infty)  \Big) = 0.
	\end{equation*}
	Therefore, the dynamics lead to sampling from a given distribution $p_{\mathrm{target}}(\rvx)$.

	\subsection{Validity of Field Divergences.}\label{sec:validity_field_divergence}
	The first step is to check whether  $\mathcal{D}_{\mathrm{grad}}$, $\mathcal{D}_{\mathrm{id}}$ are well defined divergence measures. To this end, we have the below conclusion.

	\begin{theorem}[Well-defined Divergences] \label{theorem:measure def}
		Suppose that $p(\cdot), q(\cdot)$ are probability densities that are second-order continuously differentiable (i.e., in $\mathcal C^2$) and that $p(\rvx)\neq 0$ for all $\rvx$. Then the divergence measure $\mathcal{D}_{\mathrm{grad}}({p(\cdot)}, q(\cdot))$ defined by Eq.~(\ref{eq:def of grad div}) and that $\mathcal{D}_{\mathrm{id}}({p(\cdot)}, q(\cdot))$ formulated by Eq.~(\ref{eq:def of scalar div}) are both valid statistical divergence measures, satisfying the following three conditions:
		\begin{itemize}
			\item Non-negativity: $\mathcal{D}_{*}(p(\cdot), q(\cdot))$ is either zero or positive;
			\item Null condition: $\mathcal{D}_{*}(p(\cdot), q(\cdot)) = 0$ if and only if $p(\rvx) = q(\mathbf{x})$ for every point $\mathbf{x}$;
			\item Positive definiteness: $\mathcal{D}_{*}(p(\cdot), p(\cdot) + \delta p(\cdot))$ is a positive-definite quadratic form for any infinitesimal displacement $\delta p(\cdot)$ from $p(\cdot)$.
		\end{itemize}
		Here the subscript $*$ represents either $\mathrm{grad}$ or $\mathrm{id}$.
		\begin{proof}
			Non-negativity condition obviously holds for  both $ D_{\mathrm{grad}} $ and $ D_{\mathrm{id}} $, due to their definitions.
			
			For the null condition, 
			\[
			D_{\mathrm{grad}}(p(\cdot), q(\cdot)) = 0 \quad \text{implies} \quad p(\rvx)\|\nabla p(\rvx) - \nabla q(\rvx)\|^2 = 0 \text{ for all } \rvx.
			\]

			This implies $ \nabla p(\rvx) = \nabla q(\rvx) $ for all $ x $. Since the gradients are equal, $ p(\rvx) $ and $ q(\rvx) $ differ by at most a constant. For probability densities, this constant must be zero, so $ p(\rvx) = q(\rvx) $. On the other hand, for $ D_{\mathrm{id}}(p(\cdot), q(\cdot)) = 0 $:
			\[
			D_{\mathrm{id}}(p(\cdot), q(\cdot)) = 0 \quad \text{implies} \quad (p(\rvx) - q(\rvx))^2 = 0 \text{ for all } \rvx
			\]
			This directly implies that $ p(\rvx) = q(\rvx) $ for all $ \rvx $. Hence, both $ D_{\mathrm{grad}} $ and $ D_{\mathrm{id}} $ satisfy the null condition: $ D_*(p(\cdot), q(\cdot)) = 0 $ if and only if $ p(\rvx) = q(\rvx) $ for all $ \rvx $.
			
			At last, we prove that the two measurements satisfy the positive definiteness condition. For $ D_{\mathrm{grad}}(p(\cdot), p(\cdot) + \delta p(\cdot)) $:
			\[
			D_{\mathrm{grad}}(p(\cdot), p(\cdot) + \delta p(\cdot)) = \int p(\rvx) \|\nabla p(\rvx) - \nabla(p(\rvx) + \delta p(\rvx))\|^2 \, \diff\rvx = \int p(\rvx) \|\nabla \delta p(\rvx)\|^2 \, \diff\rvx.
			\]
			
			This expression is quadratic in $ \delta p(\rvx) $, and since norms are positive definite, $ D_{\mathrm{grad}} $ is positive definite for any infinitesimal displacement $ \delta p(\rvx) $. On the other hand, for $ D_{\mathrm{id}}(p(\cdot), p(\cdot) + \delta p(\cdot)) $:
			\[
			D_{\mathrm{id}}(p(\cdot), p(\cdot) + \delta p(\cdot)) = \int p(\rvx) (p(\rvx) - (p(\rvx) + \delta p(\rvx)))^2 \, \diff\rvx = \int p(\rvx) (\delta p(\rvx))^2 \, \diff\rvx.
			\]
			Again, this is quadratic in $ \delta p(\rvx) $, making $ D_{\mathrm{id}} $ positive definite for any infinitesimal displacement $ \delta p(\rvx) $. Thus, both $ D_{\mathrm{grad}} $ and $ D_{\mathrm{id}} $ are positive-definite quadratic forms for any infinitesimal displacement $ \delta p(\rvx) $ from $ p(\rvx) $. This concludes the proof.
		\end{proof}
	\end{theorem}
	
	Since measures $\mathcal{D}_{\mathrm{grad}}$, $\mathcal{D}_{\mathrm{id}}$ are well defined, it is valid to derive the corresponding loss functions $\mathcal{L}_{\mathrm{grad}}, \mathcal{L}_{\mathrm{id}}$ as formulated in Eq.~(\ref{eq:initial loss forms}).
	
	\subsection{Proof to Proposition~\ref{prop:reshaped loss}.}
	We prove for the case of 
	\begin{equation*}
		\mathcal{L}_{\mathrm{id}}(\bm{\varphi}) = C_{\mathrm{id}} + \lim_{\epsilon \rightarrow 0} \mathbb{E}_{\mathbf{x}_1, \mathbf{x}_2 \sim {p_{\mathrm{target}}(\rvx)}}\Big[s_{\bm{\varphi}}(\mathbf{x}_1)^2  - 2 s_{\bm{\varphi}}(\mathbf{x}_1) \mathcal{N}(\mathbf{x}_2 - \mathbf{x}_1; \mathbf{0}, \epsilon \mathbf{I}_D)\Big].
	\end{equation*}
	Similar argument can be applied to the case for $\mathcal{L}_{\mathrm{grad}}(\bm{\phi})$. 
	Suppose that we have a divergence loss function:
	\begin{equation*}
		\mathcal{L}_{\mathrm{id}}(\bm{\varphi})  =  \int p_{\mathrm{target}}(\mathbf{x}) \Big( p_{\mathrm{target}}(\mathbf{x}) - s_{\bm{\varphi}}(\mathbf{x}) \Big)^2 \diff\mathbf{x}.
	\end{equation*}
	Then, we can expand the term as
	\begin{equation*}
		\mathcal{L}_{\mathrm{id}}(\bm{\varphi})  =   \int p_{\mathrm{target}}(\mathbf{x})^3 \diff\mathbf{x} - 2 \int p_{\mathrm{target}}(\mathbf{x})^2 s_{\bm{\varphi}}(\mathbf{x}) \diff\mathbf{x} + \int p_{\mathrm{target}}(\mathbf{x}) s_{\bm{\varphi}}(\mathbf{x})^2 \diff\mathbf{x}.
	\end{equation*}
	For the second integral, we apply the trick again:
	\begin{equation*}
		\begin{aligned}
			& \int p_{\mathrm{target}}(\mathbf{x})^2 s_{\bm{\varphi}}(\mathbf{x}) \diff\mathbf{x}  = \int p_{\mathrm{target}}(\mathbf{x}) p_{\mathrm{target}}(\mathbf{y}) s_{\bm{\varphi}}(\mathbf{x}) \delta(\mathbf{y} - \mathbf{x}) \diff\mathbf{x} \diff\mathbf{y} \\
			=  &\mathbb{E}_{\mathbf{x} \sim p_{\mathrm{target}}(\mathbf{x}), \mathbf{y} \sim p_{\mathrm{target}}(\mathbf{y})} \Big[ s_{\bm{\varphi}}(\mathbf{x}) \delta(\mathbf{y} - \mathbf{x}) \Big] \\
			= &\lim_{\epsilon \rightarrow 0} \mathbb{E}_{\mathbf{x}_1, \mathbf{x}_2 \sim p_{\mathrm{target}}(\mathbf{x})} \Big[s_{\bm{\varphi}}(\mathbf{x}_1) \mathcal{N}(\mathbf{x}_2 - \mathbf{x}_1; \mathbf{0}, \epsilon \mathbf{I}_D) \Big].
		\end{aligned}
	\end{equation*}
	Here, we use that $\mathcal N (\cdot; 0, \epsilon\mathbf{I}_D)$ weakly converges to $\delta(\cdot)$ as $\epsilon\rightarrow0$. Therefore, we simplify the loss function as
	\begin{equation*}
		\begin{aligned}
			\mathcal{L}_{\mathrm{id}}(\bm{\varphi}) & = C_{\mathrm{id}} + \mathbb{E}_{\mathbf{x} \sim p_{\mathrm{target}}(\mathbf{x})} \Big[s_{\bm{\varphi}}(\mathbf{x})^2 \Big] -2 \lim_{\epsilon \rightarrow 0} \mathbb{E}_{\mathbf{x}_1, \mathbf{x}_2 \sim p_{\mathrm{target}}(\mathbf{x})} \Big[ s_{\bm{\varphi}}(\mathbf{x}_1) \mathcal{N}(\mathbf{x}_2 - \mathbf{x}_1; \mathbf{0}, \epsilon \mathbf{I}_D) \Big],
		\end{aligned}
	\end{equation*}
	where $C_{\mathrm{id}}$ is a constant without learnable parameter $\bm{\varphi}$.
	
	\subsection{Proof to Proposition~\ref{prop:sliced gradient matching}}
	
	\begin{proof}
		We recall the sliced version of  $\mathcal{L}_{\mathrm{grad}}$ as:
		\begin{equation*}
			\mathcal{L}_{\mathrm{grad}}^{\mathrm{slice}}(\bm{\phi}) = \mathbb{E}_{\mathbf{v} \sim q(\mathbf{v}), \mathbf{x} \sim {p_{\mathrm{target}}(\rvx)}} \Big[ \big( \mathbf{v}^T \nabla {p_{\mathrm{target}}(\rvx)} - \mathbf{v}^T \mathbf{g}_{\bm{\phi}}(\mathbf{x})  \big)^2 \Big].
		\end{equation*}    
		
		By expanding the quadratic term $(\cdot)^2$ inside the recursive expectations, we have
		\begin{align*}
			\mathcal{L}_{\mathrm{grad}}^{\mathrm{slice}} = \mathbb{E}_{\mathbf{v}} \Big[  \int p_{\mathrm{target}}(\mathbf{x}) (\mathbf{v}^{\top} \nabla_{\mathbf{x}} \mathbf{g}_{\bm{\phi}}(\mathbf{x}))^2 \diff\rvx  -2 \int p_{\mathrm{target}}(\mathbf{x}) (\mathbf{v}^{\top} \nabla_{\mathbf{x}} p_{\mathrm{target}}(\mathbf{x}) ) (\mathbf{v}^{\top} \nabla_{\mathbf{x}} \mathbf{g}_{\bm{\phi}}(\mathbf{x})) \diff\mathbf{x} + C_{\mathrm{grad}}' \Big],
		\end{align*}
		where, $C_{\mathrm{grad}}'  :=  \int p_{\mathrm{target}}(\mathbf{x}) (\mathbf{v}^{\top} \nabla_{\mathbf{x}} p_{\mathrm{target}}(\mathbf{x}) )^2  \diff\mathbf{x} $ is a constant independent of trainable parameter.
		We will further simplify the second term came from the cross product as:
		\begin{equation*}
			\begin{aligned}
				& \int p_{\mathrm{target}}(\mathbf{x}) (\mathbf{v}^{\top} \nabla_{\mathbf{x}} p_{\mathrm{target}}(\mathbf{x}) ) (\mathbf{v}^{\top} \nabla_{\mathbf{x}} \mathbf{g}_{\bm{\phi}}(\mathbf{x})) \diff\rvx  \\
				= & \frac{1}{2} \int (\mathbf{v}^{\top} \nabla_{\mathbf{x}} p_{\mathrm{target}}(\mathbf{x})^2) (\mathbf{v}^{\top} \nabla_{\mathbf{x}} \mathbf{g}_{\bm{\phi}}(\mathbf{x})) \diff\rvx \\
				= &\frac{1}{2} \int \Big(\nabla_{\mathbf{x}} p_{\mathrm{target}}(\mathbf{x})^2\Big)^{\top} \Big( (\mathbf{v}^{\top} \nabla_{\mathbf{x}} \mathbf{g}_{\bm{\phi}}(\mathbf{x})) \mathbf{v} \Big) \diff\rvx.
			\end{aligned}
		\end{equation*}
		Note that we can replace the gradient field $\nabla_{\mathbf{x}} \mathbf{g}_{\bm{\phi}}(\mathbf{x})$ with neural network $\mathbf{g}_{\bm{\phi}}(\mathbf{x})$. By applying the integration by parts, this equality can be expanded as
		\begin{equation*}
			\frac{1}{2} \int \nabla_{\mathbf{x}} \cdot \Big( p_{\mathrm{target}}(\mathbf{x})^2 (\mathbf{v}^{\top} \mathbf{g}_{\bm{\phi}}(\mathbf{x})) \mathbf{v}  \Big) \diff\rvx -\frac{1}{2} \int p_{\mathrm{target}}(\mathbf{x})^2 \nabla_{\mathbf{x}} \cdot \Big( (\mathbf{v}^{\top} \mathbf{g}_{\bm{\phi}}(\mathbf{x})) \mathbf{v}  \Big) \diff\rvx.
		\end{equation*}
		Let us first handle the first term in the above equation. Applying Gauss's divergence theorem to a ball $\mathbb{B}(R)$ centered at the origin with radius $R>0$, we get
		\begin{equation}
			\int_{\mathbb{B}(R)} \nabla_{\mathbf{x}} \cdot \Big( p_{\mathrm{target}}(\mathbf{x})^2 (\mathbf{v}^{\top} \mathbf{g}_{\bm{\phi}}(\mathbf{x})) \mathbf{v}  \Big) \diff\rvx = \int_{\partial \mathbb{B}(R)} \mathbf{n}(\mathbf{x})^{\top} \Big( p_{\mathrm{target}}(\mathbf{x})^2 (\mathbf{v}^{\top} \mathbf{g}_{\bm{\phi}}(\mathbf{x})) \mathbf{v}  \Big).
		\end{equation}
		where $\mathbf{n}(\mathbf{x})$ is the unit norm vector to the region boundary $\partial \mathbb{B}(R)$. Suppose that $p_{\mathrm{target}}(\mathbf{x})$ decays sufficiently fast as $\norm{\mathbf{x}}_2 \rightarrow \infty$,
		for instance, $\lim_{\|\mathbf{x}\|_2 \rightarrow \infty} p_{\mathrm{target}}(\mathbf{x}) / \norm{\rvx}_2^{D} = 0$ (see Assumption~\ref{assumption_1} (iii)), then this term vanishes as $R \rightarrow \infty$.
		
		For the second term in the expansion, we have
		\begin{equation*}
			\begin{aligned}
				\nabla_{\mathbf{x}} \cdot \Big( (\mathbf{v}^{\top} \mathbf{g}_{\bm{\phi}}(\mathbf{x})) \mathbf{v}  \Big) = \sum_{1 \le i \le D} \frac{\partial ((\mathbf{v}^{\top} \mathbf{g}_{\bm{\phi}}(\mathbf{x})) v_i)}{\partial x_i} 
				= \sum_{1 \le i \le D} \sum_{1 \le j \le D} \frac{v_i v_j \rvg_{\bm{\phi}, j}(\mathbf{x})}{\partial x_i} = \mathbf{v}^{\top} \nabla_{\mathbf{x}} \rvg_{\bm{\phi}}(\mathbf{x}) \mathbf{v}.
			\end{aligned}
		\end{equation*}
		Here, we write $\rvv = (v_i)_{1\le i \le D}$ and $\rvx = (x_i)_{1\le i \le D}$.
		Collecting the above derivations, we have
		\begin{equation}
			\int p_{\mathrm{target}}(\mathbf{x}) (\mathbf{v}^{\top} \nabla_{\mathbf{x}} p_{\mathrm{target}}(\mathbf{x}) ) (\mathbf{v}^{\top} \nabla_{\mathbf{x}} \mathbf{g}_{\bm{\phi}}(\mathbf{x})) \diff\rvx = - \frac{1}{2} \int p_{\mathrm{target}}(\mathbf{x})^2 (\mathbf{v}^{\top} \nabla_{\mathbf{x}} \rvg_{\bm{\phi}}(\mathbf{x}) \mathbf{v}) \diff\rvx.
		\end{equation}
		Therefore, the loss function can be converted into
		\begin{equation}
			\mathcal{L}_{\mathrm{grad}}^{\mathrm{slice}} = \mathbb{E}_{\mathbf{v}} \Big[  \int p_{\mathrm{target}}(\mathbf{x}) (\mathbf{v}^{\top} \nabla_{\mathbf{x}} \mathbf{g}_{\bm{\phi}}(\mathbf{x}))^2 \diff\rvx  + \int p_{\mathrm{target}}(\mathbf{x})^2 (\mathbf{v}^{\top} \nabla_{\mathbf{x}} g_{\bm{\theta}}(\mathbf{x}) \mathbf{v}) \diff\rvx \Big]+ C_{\mathrm{grad}}'.
		\end{equation}
		We apply the same trick from the proof of Proposition~\ref{prop:reshaped loss}—using Dirac expansion—to enable Monte Carlo estimation for the second inner term:
		\begin{equation*}
			\begin{aligned}
				& \int p_{\mathrm{target}}(\mathbf{x})^2 \Big( \mathbf{v}^{\top} \nabla_{\mathbf{x}} g_{\bm{\theta}}(\mathbf{x}) \mathbf{v} \Big) \diff\rvx \\
				= & \int p_{\mathrm{target}}(\mathbf{x}) \Big(  \int p_{\mathrm{target}}(\mathbf{y}) \delta(\mathbf{y} - \mathbf{x}) d\mathbf{y} \Big) \Big( \mathbf{v}^{\top} \nabla_{\mathbf{x}} g_{\bm{\theta}}(\mathbf{x}) \mathbf{v} \Big) \diff\rvx \\
				= &\int p_{\mathrm{target}}(\mathbf{x}_1) p_{\mathrm{target}}(\mathbf{x}_2) ( \mathbf{v}^{\top} \nabla_{\mathbf{x}_1} g_{\bm{\theta}}(\mathbf{x}_1) \mathbf{v} ) \delta(\mathbf{x}_2 - \mathbf{x}_1)  \diff\rvx_1 \diff\rvx_2 \\
				= &\mathbb{E}_{\mathbf{x}_1, \mathbf{x}_2 \sim p_{\mathrm{target}}(\mathbf{x})} \Big[ ( \mathbf{v}^{\top} \nabla_{\mathbf{x}_1} g_{\bm{\theta}}(\mathbf{x}_1) \mathbf{v} ) \delta(\mathbf{x}_2 - \mathbf{x}_1) \Big].
			\end{aligned}
		\end{equation*}
		Combining the above two identities, we have
		\begin{equation}
			\mathcal{L}_{\mathrm{grad}}^{\mathrm{slice}} = \mathbb{E}_{\mathbf{v} \sim p_{\mathrm{slice}}(\mathbf{v}), \mathbf{x}_1, \mathbf{x}_2 \sim p_{\mathrm{target}}(\mathbf{x})} \Big[ (\mathbf{v}^{\top} g_{\bm{\theta}}(\mathbf{x}_1))^2 + ( \mathbf{v}^{\top} \nabla_{\mathbf{x}_1} g_{\bm{\theta}}(\mathbf{x}_1) \mathbf{v} ) \delta(\mathbf{x}_2 - \mathbf{x}_1) \Big]+ C_{\mathrm{grad}}',
		\end{equation}
		which completes the proof.
	\end{proof}
	
	\section{Proofs for Sec.~\ref{sec:theory}}\label{sec:main_proofs}
	\subsection{Prerequisites for Theoretical Analysis.}
	We introduce some notations and terminologies. We recall the definition of \emph{KL divergence} between $p_{\mathrm{target}}$ and density $p$ as 
	\begin{equation*}
		\mathrm{KL}\big(p\Vert p_{\mathrm{target}}\big):=\int_{\mathbb{R}^D} p(\rvx) \log \frac{p(\rvx)}{p_{\mathrm{target}}(\rvx)} \diff \rvx.
	\end{equation*}
	\emph{Fisher divergence} between $p_{\mathrm{target}}$ and $p$ is defined as:
	\begin{equation*}
		J_{p_{\mathrm{target}}}(p):=\int_{\mathbb{R}^D} p(\rvx) \norm{\nabla_x \log \frac{p(\rvx)}{p_{\mathrm{target}}(\rvx)}}^2 \diff \rvx.
	\end{equation*}
	\emph{Wasserstein-2 distance} ($W_2$) between $p_{\mathrm{target}}$ and $p$ is defined as:
	\begin{equation*}
		W_2^2(p, p_{\mathrm{target}}):=\inf_{\gamma\sim\Gamma(\mu,\nu)} \mathbb{E}_{(\rvx, \rvy) \sim\gamma}\norm{\rvx-\rvy}_2^2,
	\end{equation*}
	where $\Gamma(\mu,\nu)$ is the set of all couplings of $(\mu,\nu)$.
	
	The following summarizes the two assumptions for our main theorems in Sec.~\ref{sec:theory}.
	
	\begin{assumption}\label{assumption_1}
		Assume the target density $p_{\mathrm{target}}$ satisfies the following conditions:
		\begin{enumerate}[(i)]
			\item $p_{\mathrm{target}}(\cdot)\in\mathcal C^2$. That is, it is second-order continuously differentiable;  
			\item  \emph{Log-Sobolev inequality}: there is a constant $\alpha>0$ so that the following inequality holds for all continuously differentiable density $p$:  
			\begin{equation}
				\mathrm{KL}\big(p\Vert p_{\mathrm{target}}\big)\leq \frac{1}{2\alpha} J_{p_{\mathrm{target}}}(p).
			\end{equation}
			\item  $p_{\mathrm{target}}$ is either compactly supported with $M:=\norm{p_{\mathrm{target}}}_{L^{\infty}} <\infty$, or it decays sufficiently fast as $\norm{\mathbf{x}}_2 \rightarrow \infty$: 
			\begin{align*}
				\lim_{\|\mathbf{x}\|_2 \rightarrow \infty} \frac{p_{\mathrm{target}}(\mathbf{x})}{\norm{\rvx}_2^{D}} = 0.
			\end{align*}
		\end{enumerate}
	\end{assumption}
	
	\begin{assumption}\label{assumption_2}
		Assume the target density $p_{\mathrm{target}}$ satisfies the following additional conditions:
		\begin{enumerate}[(i)]
			\item There is a $L>0$ so that for all $\rvx, \rvy$
			\begin{align*}
				\norm{p_{\mathrm{target}}(\rvx) - p_{\mathrm{target}}(\rvy)}_2^2\leq L\norm{\rvx - \rvy}_2^2 \quad \text{and}\quad \norm{\nabla p_{\mathrm{target}}(\rvx) - \nabla p_{\mathrm{target}}(\rvy)}_2^2\leq L\norm{\rvx - \rvy}_2^2.
			\end{align*}
		\end{enumerate}
	\end{assumption}

	\subsection{Proofs of Theorem~\ref{thm:stationary_dist}}
	\begin{proof}
		Recall our dynamics is
		\begin{align*}
			\diff \rvx(t) = \nabla p_{\mathrm{target}}(\rvx(t)) \diff t+\sqrt{p_{\mathrm{target}}(\rvx(t))}\diff \rvw(t).
		\end{align*}
		
		The Fokker-Planck equation of our dynamics with density $p_t=p_t(\rvx):=p(\rvx,t)$ is
		\begin{equation}\label{eq:our_fp}
			\partial_t p(\rvx,t) =\frac{1}{2}\nabla\cdot \Big(p_{\mathrm{target}}(\rvx) p(\rvx,t) \nabla \log \frac{p(\rvx,t)}{p_{\mathrm{target}}(\rvx)}\Big).
		\end{equation}
		
		This is due to the following derivation \cite{risken1996fokker}, where we demonstrated for the $D=1$ case. The probability distribution $p(\rvx, t)$ of dynamics $\rvx(t)$ at point $\rvx$ and time $t$ with $f(x) = \partial_x p_{\mathrm{target}}(x)$ and $g^2(x) = p_{\mathrm{target}}(x)$ is governed by:
		\begin{align*}
			\frac{\partial p(x, t)}{\partial t} & = - \frac{\partial}{\partial x} \Big( f(x) p(x, t)  \Big) + \frac{1}{2} \frac{\partial^2}{\partial x^2} \Big( g^2(x) p(x, t)  \Big) 
			\\
			& = \frac{\partial}{\partial x}\Big( -\partial_x p_{\mathrm{target}}(x) p(x, t) + \frac{1}{2} \frac{\partial }{\partial x}\Big( p_{\mathrm{target}}(x) p(x, t) \Big) \Big)
			\\
			& = \frac{1}{2} \frac{\partial}{\partial x}\Big(p_{\mathrm{target}}(x) \partial_x p(x,t) -\partial_x p_{\mathrm{target}}(x) p(x, t)  \Big)  \\
			& = \frac{1}{2} \frac{\partial}{\partial x}\Big(p_{\mathrm{target}}(x) p(x,t) \partial_x \log\frac{p(x, t) }{p_{\mathrm{target}}(x)}   \Big).
		\end{align*}
		Here, in the last equality we use the identity:
		\begin{align*}
			\partial_\rvx \log \frac{p(x,t)}{p_{\mathrm{target}}(x)}= \frac{p_{\mathrm{target}}(x)}{p(x,t)}\partial_\rvx\Big(\frac{p(x,t)}{p_{\mathrm{target}}(x)}\Big)^2= \frac{p_{\mathrm{target}}(x) \partial_x p(x,t) -\partial_x p_{\mathrm{target}}(x) p(x, t)}{p_{\mathrm{target}}(x)p(x,t)}.
		\end{align*}
		For a general $D$, the same computation can be carried out to derive Eq.~\eqref{eq:our_fp}.
		
		We now prove the KL bound of convergence using a similar argument motivated by \cite{vempala2019rapid}.
		\begin{align*}
			&\frac{\diff }{\diff t}\mathrm{KL}\big(p_t\Vert p_{\mathrm{target}}\big)\\
			=&~\frac{\diff }{\diff t} \int_{\mathbb{R}^D} p_t \log \frac{p_t}{p_{\mathrm{target}}} \diff \rvx  
			\\ \stackrel{\tiny{(a)}}{=}&\int_{\mathbb{R}^D} \frac{\partial }{\partial t} p_t \log \frac{p_t}{p_{\mathrm{target}}} \diff \rvx  + \int_{\mathbb{R}^D}  p_t \frac{\partial }{\partial t} \log \frac{p_t}{p_{\mathrm{target}}} \diff \rvx 
			\nonumber
			\\ \stackrel{\tiny{(b)}}{=}&\int_{\mathbb{R}^D} \frac{\partial }{\partial t} p_t \log \frac{p_t}{p_{\mathrm{target}}} \diff \rvx 
			\nonumber
			\\ \stackrel{\tiny{(c)}}{=}& \frac{1}{2} \int_{\mathbb{R}^D} \Big[\nabla_x \cdot \big(p_{\mathrm{target}} p_t  \nabla_x \log\frac{p_t}{p_{\mathrm{target}}} \big) \Big] \log \frac{p_t}{p_{\mathrm{target}}} \diff \rvx 
			\nonumber
			\\ \stackrel{\tiny{(d)}}{=}&- \int_{\mathbb{R}^D}   p_{\mathrm{target}} p_t \norm{\nabla_x  \log \frac{p_t}{p_{\mathrm{target}}}}^2 \diff \rvx 
			\\ \stackrel{\tiny{(e)}}{\leq}&- M\int_{\mathbb{R}^D}    p_t \norm{\nabla_x  \log \frac{p_t}{p_{\mathrm{target}}}}^2 \diff \rvx 
			\nonumber
			\\ = &-M J_{p_{\mathrm{target}}}(p_t)
			\\ \leq &-\frac{M}{2\alpha}\mathrm{KL}\big(p_t\Vert p_{\mathrm{target}}\big).
		\end{align*}
		Here, (a) follows from the chain rule; (b) uses the identity $\int p_t \frac{\partial }{\partial t} \log \frac{p_t}{p_{\mathrm{target}}} \diff \rvx = \int \frac{\partial }{\partial t} p_t \diff \rvx = \frac{\diff }{\diff t} \int p_t \diff \rvx = 0$; (c) follows from the Fokker-Planck Eq.~\eqref{eq:our_fp}; (d) is due to integration by parts and Assumption~\ref{assumption_1} (iii); and (e) comes from Assumption~\ref{assumption_1} (iii).
		
		Thus, applying Grönwall's inequality, we can get
		\begin{align*}
			\mathrm{KL}\big(p_t\Vert p_{\mathrm{target}}\big) \lesssim e^{-2\alpha t}\mathrm{KL}\big(p_0\Vert p_{\mathrm{target}}\big).
		\end{align*}
		
		Since $p_{\mathrm{target}}$ satisfies the LSI, it also satisfies the Talagrand's inequality~\cite{otto2000generalization}:
		\begin{align*}
			\frac{\alpha}{2}W_2^2\big(p_t, p_{\mathrm{target}}\big)\leq\mathrm{KL}\big(p_t\Vert p_{\mathrm{target}}\big).
		\end{align*}
		Therefore, we have
		\begin{align*}
			W_2^2\big(p_t, p_{\mathrm{target}}\big)\leq\frac{2}{\alpha}\mathrm{KL}\big(p_t\Vert p_{\mathrm{target}}\big) \lesssim\frac{2}{\alpha}e^{-2\alpha t}\mathrm{KL}\big(p_0\Vert p_{\mathrm{target}}\big).
		\end{align*}
		This completes the proof. We notice that ``Talagrand’s inequality implies concentration of measure of Gaussian type'' allowing us to remove the compact support assumption on $p_{\mathrm{target}}$ while maintaining the validity of the theorem.
	\end{proof}
	
	\subsection{Proofs of Theorem~\ref{thm:informal_error}}
	
	\begin{proof} In the proof we will extensively using a simple form of Cauchy-Schwarz (CS) inequality:
		\begin{align*}
			(u_1+u_2+\cdots+u_n)^2 \leq n (u_1^2+u_2^2+\cdots+u_n^2),
		\end{align*}
		for $u_i\in\mathbb{R}$, $i=1,\cdots,n$. We aim at obtaining the following bound:
		\begin{align}\label{eq:precise_error_bound}
			W_2^2(p_{T;\bm{\phi},\bm{\varphi}}, p_{\mathrm{target}})\lesssim \varepsilon_{\mathrm{est}}^2 T e^{LT} + \frac{2}{\alpha}e^{-\alpha T}\mathrm{KL}\big(p_0\Vert p_{\mathrm{target}}\big).
		\end{align}
		To achieve it, we compare the random vector processes $\{\mathbf{x}(t)\}_{t\in[0,T]}$ and $\{\widehat{\rvx}(t)\}_{t\in[0,T]}$, governed by the following dynamics:
		\begin{align*}
			\diff\mathbf{x}(t) &= \nabla p_{\mathrm{target}}(\mathbf{x}(t)) \diff t + \sqrt{p_{\mathrm{target}}(\mathbf{x}(t))} \diff\bm{\omega}(t) \\
			\diff \widehat{\rvx}(t) &= g_{\bm{\phi}}(\widehat{\rvx}(t)) \diff t+\sqrt{s_{\bm{\varphi}}(\widehat{\rvx}(t))}\diff \widehat{\rvw}(t).
		\end{align*}
		Their strong solutions in the Itô sense are:
		\begin{align*}
			\mathbf{x}(t) &=  \mathbf{x}(0) + \int_{0}^{T} \nabla p_{\mathrm{target}}(\mathbf{x}(t)) \diff t + \int_{0}^{T} \sqrt{p_{\mathrm{target}}(\mathbf{x}(t))} \diff\bm{\omega}(t) \\
			\widehat{\rvx}(t) &= \widehat{\rvx}(0) + \int_{0}^{T} g_{\bm{\phi}}(\widehat{\rvx}(t)) \diff t+ \int_{0}^{T} \sqrt{s_{\bm{\varphi}}(\widehat{\rvx}(t))}\diff \widehat{\rvw}(t).
		\end{align*}
		Set random vectors $\rva(t):=\nabla p_{\mathrm{target}}(\mathbf{x}(t))  - g_{\bm{\phi}}(\widehat{\rvx}(t))$ and $\rvb(t):=\sqrt{p_{\mathrm{target}}(\mathbf{x}(t))}  - \sqrt{s_{\bm{\varphi}}(\widehat{\rvx}(t))}$, we then have
		\begin{align*}
			\mathbb{E}\big[\norm{\mathbf{x}(T)-\widehat{\rvx}(T)}_2^2\big]&\leq \mathbb{E}\Bigg[\Big(\mathbf{x}(0)-\widehat{\rvx}(0)+ \int_{0}^{T}\rva(t) \diff t + \int_{0}^{T}\rvb(t) \diff\bm{\omega}(t) \Big)^2\Bigg]  \\
			&\leq 3 \mathbb{E}\Big[\norm{\mathbf{x}(0)-\widehat{\rvx}(0)}_2^2 \Big]+ 3 \mathbb{E}\Big[\big(\int_{0}^{T}\rva(t) \diff t \big)^2 \Big] + 3 \mathbb{E}\Big[\big( \int_{0}^{T}\rvb(t) \diff\bm{\omega}(t)\big)^2 \Big]  \\
			&\lesssim \mathbb{E}\Big[\norm{\mathbf{x}(0)-\widehat{\rvx}(0)}_2^2 \Big]+ T \mathbb{E}\Big[\int_{0}^{T}\rva^2(t) \diff t \Big] + \mathbb{E}\Big[ \int_{0}^{T}\rvb^2(t) \diff t \Big]  \\
			&\lesssim \mathbb{E}\Big[\norm{\mathbf{x}(0)-\widehat{\rvx}(0)}_2^2 \Big] + T \mathbb{E}\Big[\int_{0}^{T} \norm{\nabla p_{\mathrm{target}}(\mathbf{x}(t))  - \nabla p_{\mathrm{target}}(\widehat{\rvx}(t))}_2^2 \diff t \Big] 
			\\  &~~+ T\mathbb{E}\Big[\int_{0}^{T} \norm{\nabla p_{\mathrm{target}}(\widehat{\rvx}(t)) -  g_{\bm{\phi}}(\widehat{\rvx}(t))}_2^2 \diff t \Big] \\  &~~+ \mathbb{E}\Big[ \int_{0}^{T} \abs{p_{\mathrm{target}}(\mathbf{x}(t) - p_{\mathrm{target}}(\widehat{\rvx}(t))} \diff t \Big]  + \mathbb{E}\Big[\int_{0}^{T} \abs{p_{\mathrm{target}}(\widehat{\rvx}(t)) - s_{\bm{\varphi}}(\widehat{\rvx}(t))} \diff t \Big]\\
			&\lesssim \mathbb{E}\Big[\norm{\mathbf{x}(0)-\widehat{\rvx}(0)}_2^2 \Big] +L T \int_{0}^{T} \mathbb{E}\big[\norm{\mathbf{x}(t)-\widehat{\rvx}(t)}_2^2\big]\diff t+ \varepsilon_{\mathrm{est}}^2 T.
		\end{align*}
		Here, we apply the Cauchy-Schwarz (CS) inequality and the Itô isometry in the third inequality, the CS inequality and $(\sqrt{u}-\sqrt{v})^2 \leq \abs{u-v}$ ($u$, $v\geq0$) in the fourth inequality, and the estimation error assumption in the last equality.
		
		Since the dynamics in Eqs.~\eqref{eq:new sampling method} and \eqref{eq:nn new sampling method} start from the same initial condition sampled from $p_0$, we have $\mathbb{E}\Big[\norm{\mathbf{x}(0)-\widehat{\rvx}(0)}_2^2 \Big]=0$.
		Applying the Grönwall's inequality and the definition of the Wasserstein-2 distance, then we obtain
		\begin{align*}
			W_2^2(p_{T;\bm{\phi},\bm{\varphi}}, p_T)  \lesssim \varepsilon_{\mathrm{est}}^2 T e^{LT}.
		\end{align*}
		Combining the above inequality and the result of Theorem~\ref{thm:stationary_dist} that 
		\begin{align*}
			W_2^2\big(p_T, p_{\mathrm{target}}\big)\lesssim\frac{2}{\alpha}e^{-\alpha T}\mathrm{KL}\big(p_0\Vert p_{\mathrm{target}}\big),
		\end{align*}
		we finally derive the following inequality by applying CS inequality
		\begin{align*}
			W_2^2(p_{T;\bm{\phi},\bm{\varphi}}, p_{\mathrm{target}})\lesssim \varepsilon_{\mathrm{est}}^2 T e^{LT} + \frac{2}{\alpha}e^{-\alpha T}\mathrm{KL}\big(p_0\Vert p_{\mathrm{target}}\big).
		\end{align*}
	\end{proof}
	
	\section{Bellman Diffusion for MDPs}
	\label{sec:apply to mdp}
	
	\begin{figure*}
		\centering 
		\includegraphics[width=0.7\textwidth]{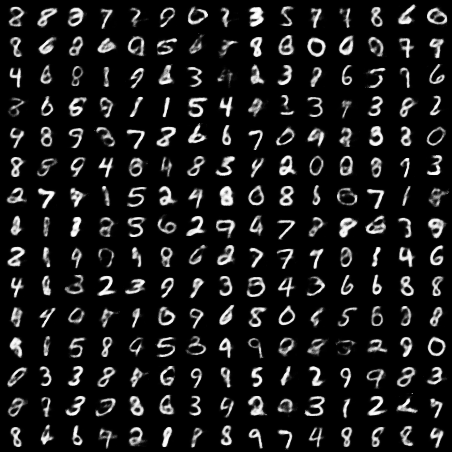}
		\caption{$15 \times 15 $ randomly sampled images from our latent Bellman Diffusion model that is trained on the MNIST dataset. We can see that most of the results are high-quality.} 
		\label{fig:mnist results}
	\end{figure*}
	
	While it is clear that Bellman Diffusion can be used as a generative model, we show that our framework applies to MDPs, which is not the case of its counterpart (e.g., SGMs).
	
	\subsection{Disfavored Full Trajectory Sampling}\label{sec:full_trajectory}
	
	As mentioned in Sec.~\ref{sec:prelim}, a DGM that is qualified to be applied with the efficient Bellman update needs to satisfy some linearity condition, otherwise one can only sample full state-action trajectories to train the DGM, which is too costly for many RL environments. To understand this point, suppose that there is an $1$-dimensional maze with $N$ blocks, with a robot moving from the leftmost block to the rightmost block. If one directly trains the return model with the returns computed from full trajectories, then the robot has to try to move to the final block after each action, resulting in a time complexity at least as $\mathcal{O}(N \cdot N) = \mathcal{O}(N^2)$ for every episode. In contrast, if the return model can be trained with partial trajectories (e.g., $1$ step) through the Bellman equation, then the time complexity would be significantly reduced (e.g., $\mathcal{O}(N^2)$). There are many RL environments where the number $N$ can be very big. For example, StarCraft II~\citep{vinyals2017starcraft} and Counter-Strike~\citep{pearce2022counter}, where a full trajectory can contain over ten thousand steps.
	
	\subsection{Efficient Bellman Diffusion Update}
	
	\begin{algorithm*}
		\caption{Planning with Bellman Diffusion} 
		\label{alg:planning algo}
		\begin{algorithmic}[1]
			\Repeat
			\State Sample state transition $(z_t, z_{t+1}, r_t)$ from the MDP and policy $\pi$
			\If{$z_t$ is the end state}
			\Comment{$z_{t+1}$ is just a dummy variable in this case.}
			\State Sample $x_1, x_2$ from $\mathcal{N}(r_t; 0, \xi)$
			\State $\mathcal{L}_{\mathrm{grad}}(\epsilon) = g_{\bm{\phi}}(x_1, z_t)^2 + \mathcal{N}(x_1 - x_2; 0, \epsilon) \partial_{x_1} g_{\bm{\phi}}(x_1, z_t)$
			\State $\mathcal{L}_{\mathrm{id}}(\epsilon) = s_{\bm{\varphi}}(x_1, z_t)^2  - 2 \mathcal{N}(x_1 - x_2; 0, \epsilon) s_{\bm{\varphi}}(x_1, z_t) $
			\State Update parameter $\bm{\phi}$ with $-\nabla_{\bm{\phi}} \mathcal{L}_{\mathrm{grad}}(\epsilon)$
			\State Update parameter $\bm{\varphi}$ with $-\nabla_{\bm{\varphi}} \mathcal{L}_{\mathrm{id}}(\epsilon)$
			\Else
			\State Sample $x$ from a bounded span $(\mathrm{x}_{\mathrm{min}}, \mathrm{x}_{\mathrm{max}})$
			\State Set target gradient $g_{\mathrm{tgt}} = g_{\bm{\phi}}\big(\frac{x - r}{\gamma}, z_{t+1}\big)$
			\State Set target scalar  $s_{\mathrm{tgt}} = \frac{1}{\gamma}s_{\bm{\varphi}}\big( \frac{x - r}{\gamma}, z_{t+1} \big)$
			\State Update parameter $\bm{\phi}$ with $ - \nabla_{\bm{\phi}} \big(g_{\bm{\phi}}(x, z_t) - g_{\mathrm{tgt}} \big)^2 $
			\State Update parameter $\bm{\varphi}$ with $ - \nabla_{\bm{\varphi}} \big(s_{\bm{\varphi}}(x, z_t) - s_{\mathrm{tgt}} \big)^2 $
			\EndIf
			\Until{parameters $\bm{\phi}, \bm{\varphi}$ converge}
		\end{algorithmic}
	\end{algorithm*}
	
	Conventionally, the MDP is specified by a $5$-tuple $(\mathcal{Z}, \mathcal{A}, p_{\mathrm{tran}}, p_{\mathrm{rwd}}, \gamma)$, where $\mathcal{Z}$ is the state space, $\mathcal{A}$ is the action space, $p_{\mathrm{tran}}: \mathcal{Z} \times \mathcal{A} \times \mathcal{Z} \rightarrow \mathbb{R}$ represents the transition probability, $p_{\mathrm{rwd}}: \mathcal{Z} \times \mathcal{A} \rightarrow \mathbb{R}$ denotes the reward model, and $\gamma$ is the discount factor. Given a policy $\pi$ that takes some action $a \in \mathcal{A}$ at every state $z \in \mathcal{Z}$, we aim to estimate the probability distribution of discounted return $X = \sum_{t \ge 1} \gamma^{t - 1} R_t$ for every state $s$ or state-action pair $(s, a)$.
	
	Let us take the state return $X_z, z \in \mathcal{Z}$ as an example. To apply Bellman Diffusion to that case, we have to parameterize the gradient and scalar field models $g_{\bm{\phi}}(x), s_{\bm{\varphi}}(x)$ for every state $z$, which is memory consuming for a large state space $\mathcal{Z}$. We adopt the notation $g_{\bm{\phi}}(x)$, rather than $\mathbf{g}_{\bm{\phi}}(x, z)$, because the return distribution is typically one dimensional. To tackle this issue, we share the model parameters $\bm{\phi}, \bm{\varphi}$ among different states, letting the models $g_{\bm{\phi}}(x, z), s_{\bm{\varphi}}(x, z)$ depend on state $z$. With this parameterization strategy, we show a planning algorithm in Algorithm~\ref{alg:planning algo}.
	
	In that procedure, we assume that the reward at the end state is a scalar and $\xi$ is a small value. This assumption applies to most of the scenarios. For example, one will win or fail in the end of a game. We also respectively set $\mathrm{x}_{\mathrm{min}}, \mathrm{x}_{\mathrm{max}}$ as the minimum and maximum returns one can get. this algorithm is for planning, and it can be further applied to RL by adding a step of action selection: most of the time choosing the action that leads to the best expected return, and otherwise doing random exploration.
	
	A baseline is C51~\citep{bellemare2017distributional}, which models the return distribution with a simple categorical distribution. Its training algorithm is as shown in Algorithm 1 of their paper. Obviously, diffusion model can model much more complex distributions than a categorical parameterization. Therefore, our model will perform better than C51 in the task of distributional RL.
	
	\section{Experiment Settings}
	\label{appendix:experiment setup}
	
	Unless specified, we construct the gradient and scalar field models $ \mathbf{g}_{\bm{\phi}}(\mathbf{x}) $ and $ s_{\bm{\varphi}}(\mathbf{x}) $ using MLPs~\citep{pinkus1999approximation}. We employ Adam algorithm~\citep{kingma2014adam} for optimization, without weight decay or dropout. The parameter $ \epsilon $ in the loss functions $\widebar{\mathcal{L}}_{\mathrm{grad}}^{\mathrm{slice}}(\bm{\phi};\epsilon)$ and $ \widebar{\mathcal{L}}_{\mathrm{id}}^{\mathrm{slice}}(\bm{\varphi};\epsilon)$ ranges from $ 0.1 $ to $ 1.0 $, depending on the task. For the sampling dynamics defined in Eq.~(\ref{eq:new sampling method}), we typically set $ T = 300 $ and $ \eta = 0.1 $. All models are trained on a single A100 GPU with 40GB memory, taking only a few tens of minutes to a few hours.
	
	\section{Additional Experiments}
	\label{appendix:extra experiments}
	
	Due to the limited space, we put the minor experiments here in the appendix. The main experiments involving field estimation, generative modeling, and RL are placed in the main text.
	
	\subsection{Image Generation}\label{sec:img_gen}
	
	While image generation is not the main focus of our paper, we show that Bellman Diffusion is also promising in that direction. We adopt a variant of the widely used architecture of latent diffusion~\citep{rombach2022high}, with VAE to encode images into latent representations and Bellman Diffusion to learn the distribution of such representations. We run such a model on MNIST~\citep{deng2012mnist}, a classical image dataset. The results are shown in Fig.~\ref{fig:mnist results}. We can see that most generated images are high-quality. This experiment verify that Bellman Diffusion is applicable to high-dimensional data, including image generation.
	
	\subsection{Ablation Studies}
	
	\begin{table*}
		\centering
		\scalebox{0.9}{     \begin{tabular}{c|cc}
				\hline
				Method & Abalone & Telemonitoring \\ \hline
				Bellman Diffusion w/ $\epsilon = 0.5, n = 1$ & $0.975$ & $2.167$ \\ \hline
				Bellman Diffusion w/ $\epsilon = 1.0, n = 1$ & $1.113$ & $2.379$ \\
				Bellman Diffusion w/ $\epsilon = 0.1, n = 1$ & $0.875$ & $2.075$ \\
				Bellman Diffusion w/ $\epsilon = 0.01, n = 1$ & $1.567$ & $3.231$ \\ \hline
				Bellman Diffusion w/ $\epsilon = 0.5, n = 2$ & $0.912$ & $2.073$ \\
				Bellman Diffusion w/ $\epsilon = 0.5, n = 3$ & $0.895$ & $1.951$ \\
				\hline
			\end{tabular}
		}
		\caption{The experiment results of our case studies, with Wasserstein distance as the metric.}
		\label{tab:ablation studies}
	\end{table*}
	
	There are some important hyper-parameters of Bellman Diffusion that need careful studies to determine their proper values for use. This part aims to achieve this goal. We adopt two tabular datasets: Abalone and Telemonitoring, with the Wasserstein distance as the metric.
	
	\paragraph{The variance of Gaussian coefficients.} The loss functions $\widebar{\mathcal{L}}_{\mathrm{grad}}(\bm{\phi};\epsilon), \widebar{\mathcal{L}}_{\mathrm{id}}(\bm{\varphi};\epsilon)$ of both gradient and scalar matching contain a term $\epsilon$, which is to relax their original limit forms for practical computation. As shown in the first $4$ rows of Table~\ref{tab:ablation studies}, either too big or too small value of term $\epsilon$ leads to worse performance of our Bellman Diffusion model. These experiment results also make sense because too big $\epsilon$ will significantly deviate the loss functions from their limit values, and too small $\epsilon$ will also cause numerical instability.
	
	\paragraph{Number of slice vectors.} Intuitively, more slice vectors will make our loss estimation more accurate, leading to better model performance. The experiment results in the first and the last two rows of Table~\ref{tab:ablation studies} confirm this intuition, but also indicate that such performance gains are not notable. Therefore, we adopt $n = 1$ slice vectors in experiments to maintain high efficiency.
	
\end{document}